\documentclass[10pt,twocolumn,letterpaper]{article}

\usepackage{cvpr}              % To produce the CAMERA-READY version
\usepackage{microtype}

\usepackage{siunitx}
\usepackage[table, dvipsnames]{xcolor}
\usepackage{comment}

\colorlet{rank1}{LimeGreen!50}
\colorlet{rank2}{SpringGreen!50}
\colorlet{rank3}{GreenYellow!40}
\colorlet{rank4}{Goldenrod!40}
\colorlet{rank5}{YellowOrange!40}
\colorlet{rank6}{Orange!40}
\colorlet{rank7}{Orange!50!Red!40}
\colorlet{rank8}{Red!45}

\usepackage[accsupp]{axessibility}  % Improves PDF readability for those with disabilities.

\definecolor{cvprblue}{rgb}{0.21,0.49,0.74}
\usepackage[pagebackref,breaklinks,colorlinks,allcolors=cvprblue]{hyperref}

\def\paperID{17} % *** Enter the Paper ID here
\def\confName{CVPR} % CVPR Workshop 3DMV
\def\confYear{2026}

\title{Indoor Asset Detection in Large Scale 360$^\circ$ Drone-Captured Imagery via 3D Gaussian Splatting}

\author{
Monica Tang \qquad Avideh Zakhor \\
UC Berkeley \\
{\tt\small \{m.tang, avz\}@berkeley.edu}
}

\begin{document}
% \maketitle
\twocolumn[{%
\renewcommand\twocolumn[1][]{#1}%
\maketitle
\begin{center}
    \vspace{-1em}
    \centering
    \begin{minipage}[H]{0.40\textwidth}
    \begin{minipage}[l]{0.95\linewidth}
        \footnotesize Input Image Views
        \vspace{0.25em}
    \end{minipage}
    \begin{minipage}[c]{0.95\linewidth}
        \centering
        \includegraphics[width=0.237\linewidth]{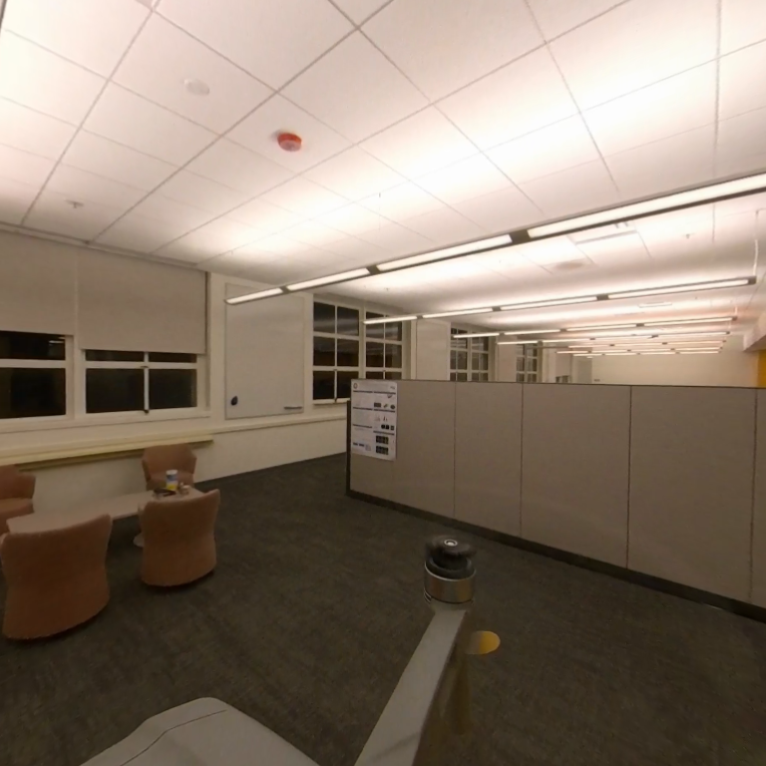}
        \includegraphics[width=0.237\linewidth]{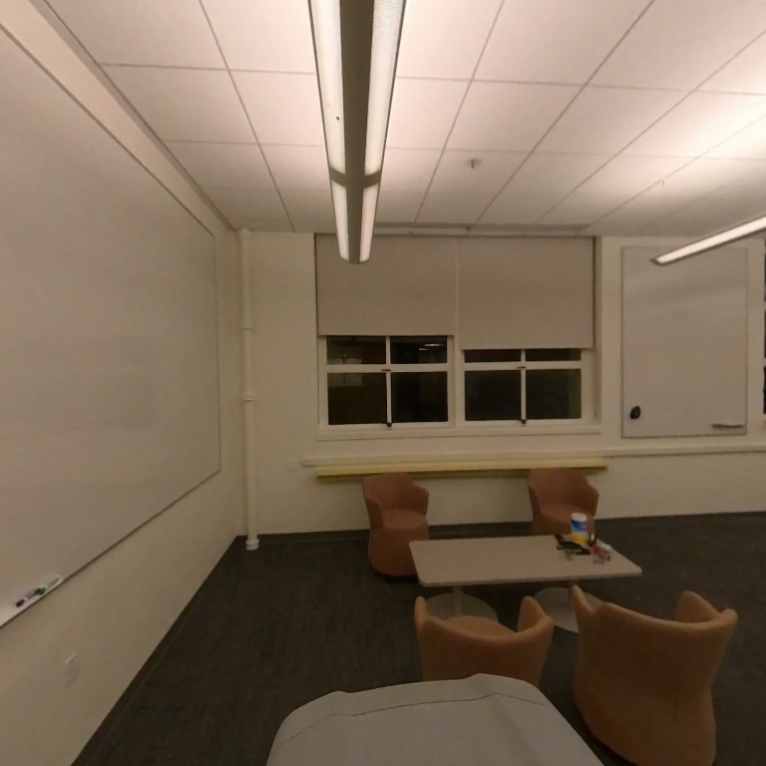}
        \includegraphics[width=0.237\linewidth]{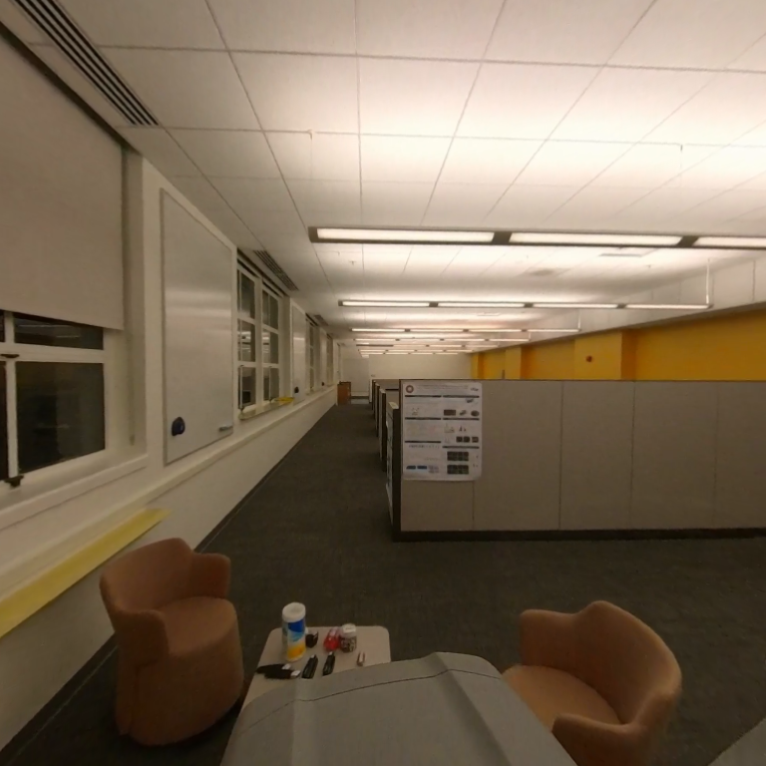}
        \includegraphics[width=0.237\linewidth]{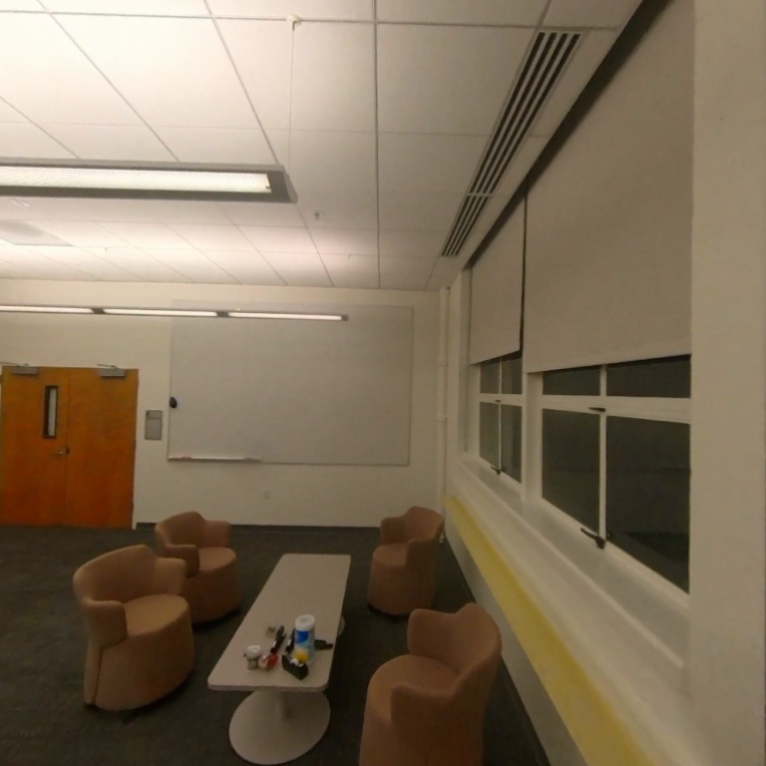}
    \end{minipage}
    \noindent 
    \\ \\
    \begin{minipage}[c]{0.95\linewidth}
        \centering
        \big \downarrow
        \vspace{0.25em}
    \end{minipage}
    \begin{minipage}[l]{0.95\linewidth}
        \footnotesize Multi-View Inconsistent Masks
        \vspace{0.25em}
    \end{minipage}
    \begin{minipage}[c]{0.95\linewidth}
        \centering
        \includegraphics[width=0.237\linewidth]{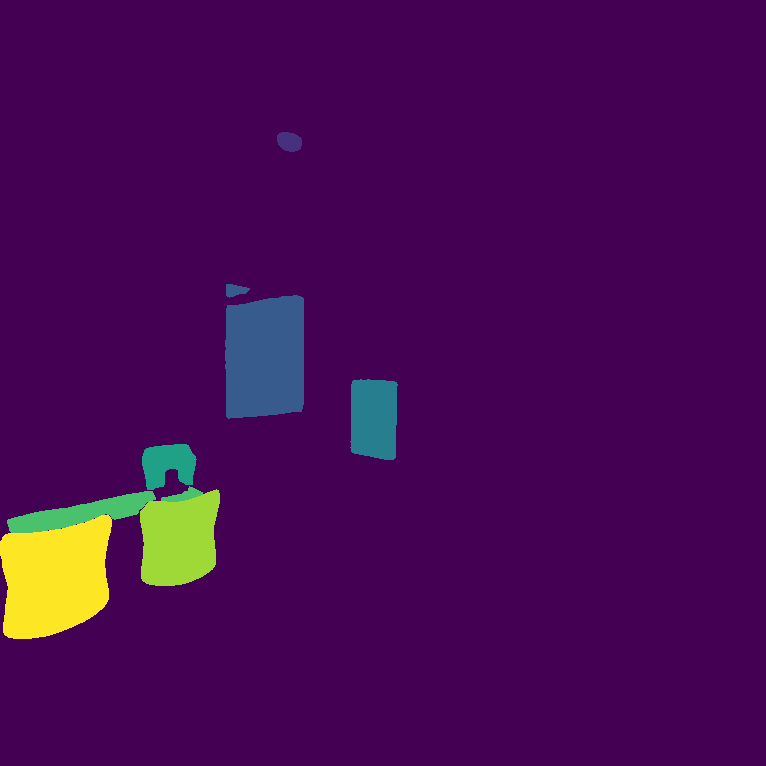}
        \includegraphics[width=0.237\linewidth]{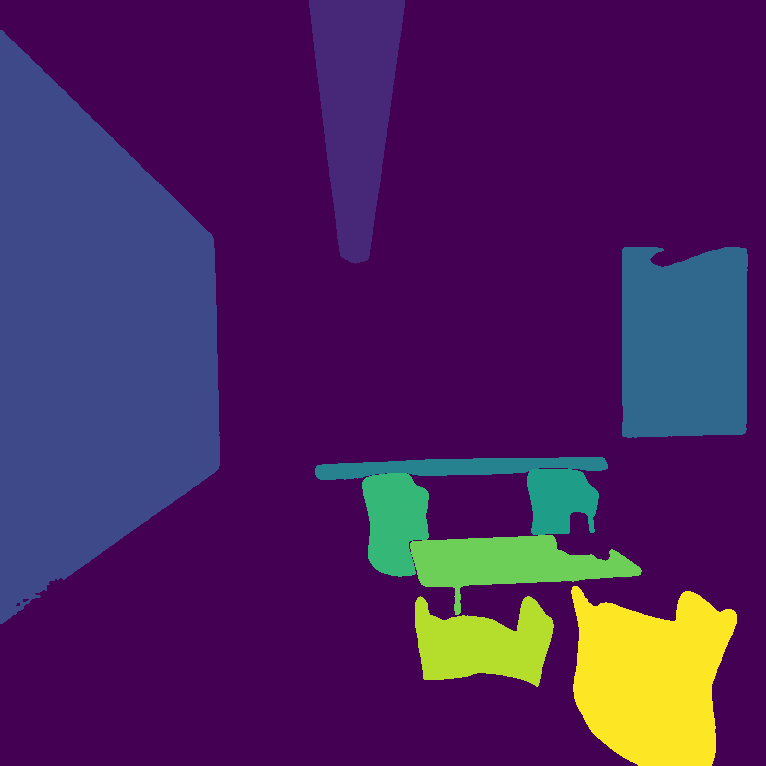}
        \includegraphics[width=0.237\linewidth]{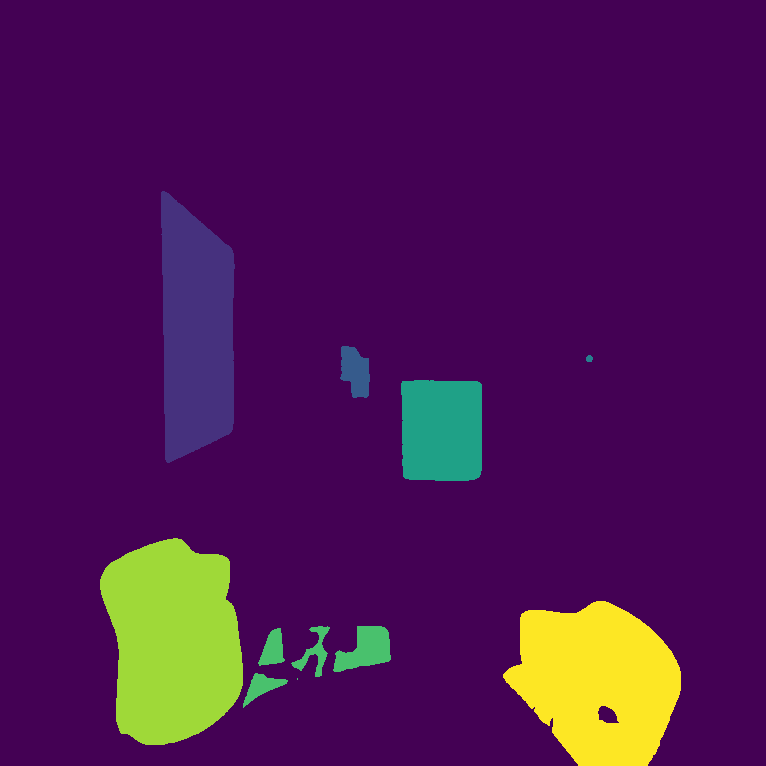}
        \includegraphics[width=0.237\linewidth]{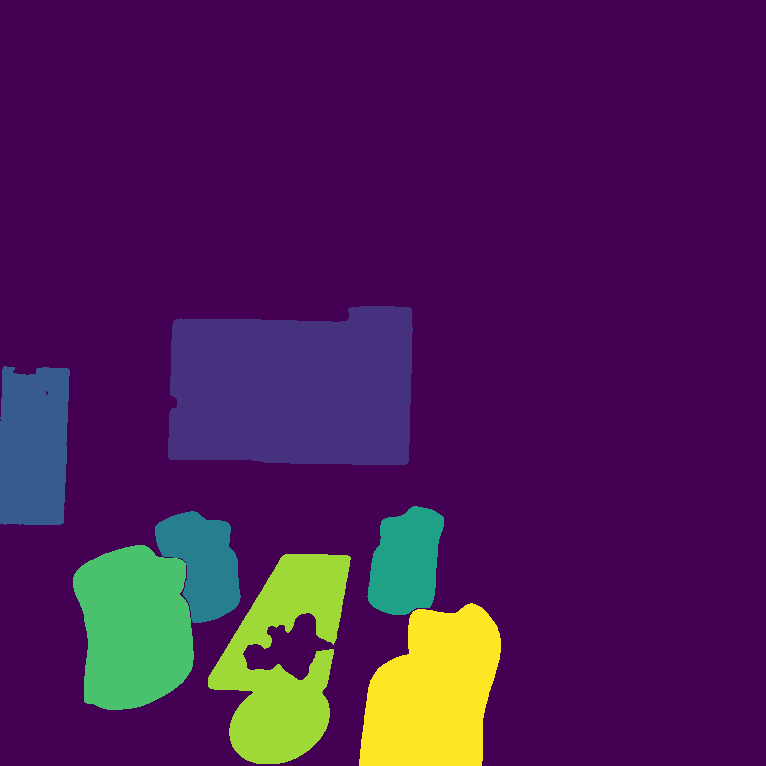}
    \end{minipage} %
    \begin{minipage}[l]{0.03\linewidth}
        $\rightarrow$
    \end{minipage}
    \end{minipage}
    \begin{minipage}[H]{0.15\textwidth}
        \centering
        \includegraphics[width=\linewidth]{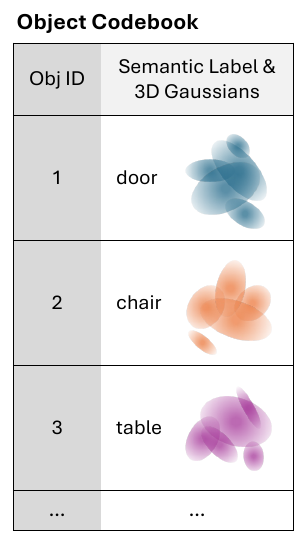}%
    \end{minipage}
    $\rightarrow$
    \begin{minipage}[H]{0.41\textwidth}
    \centering
    \begin{minipage}[c]{0.05\linewidth}
        \centering
        \rotatebox{90}{\footnotesize Ground Truth}
    \end{minipage}%
    \begin{minipage}[c]{0.95\linewidth}
        \centering
        \includegraphics[width=0.24\linewidth]{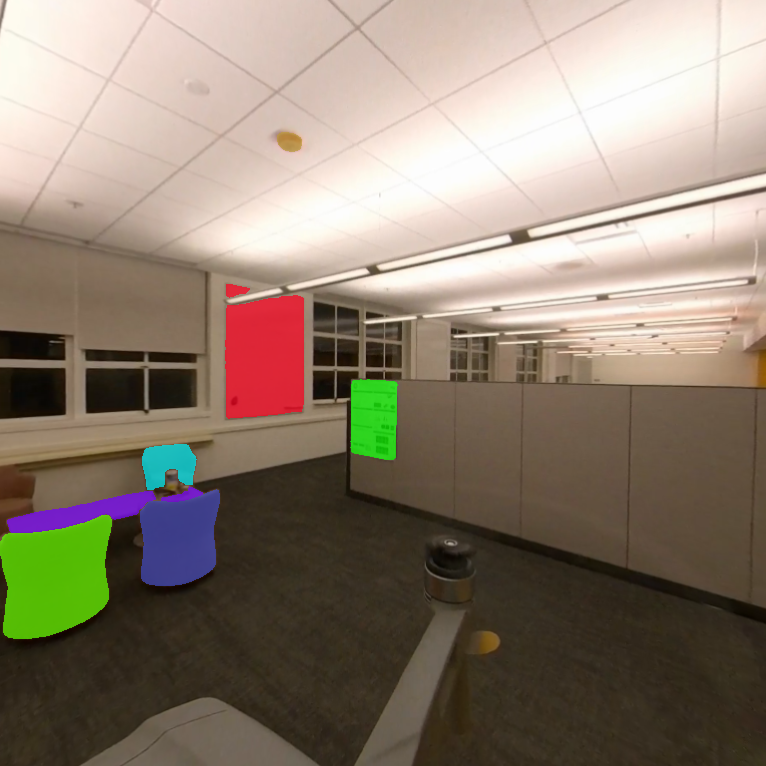}
        \includegraphics[width=0.24\linewidth]{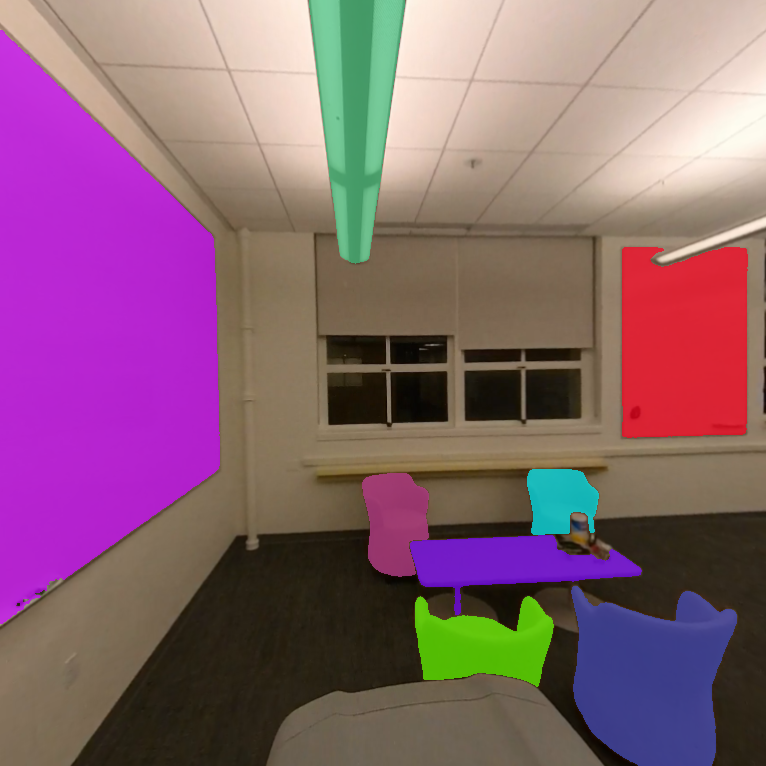}
        \includegraphics[width=0.24\linewidth]{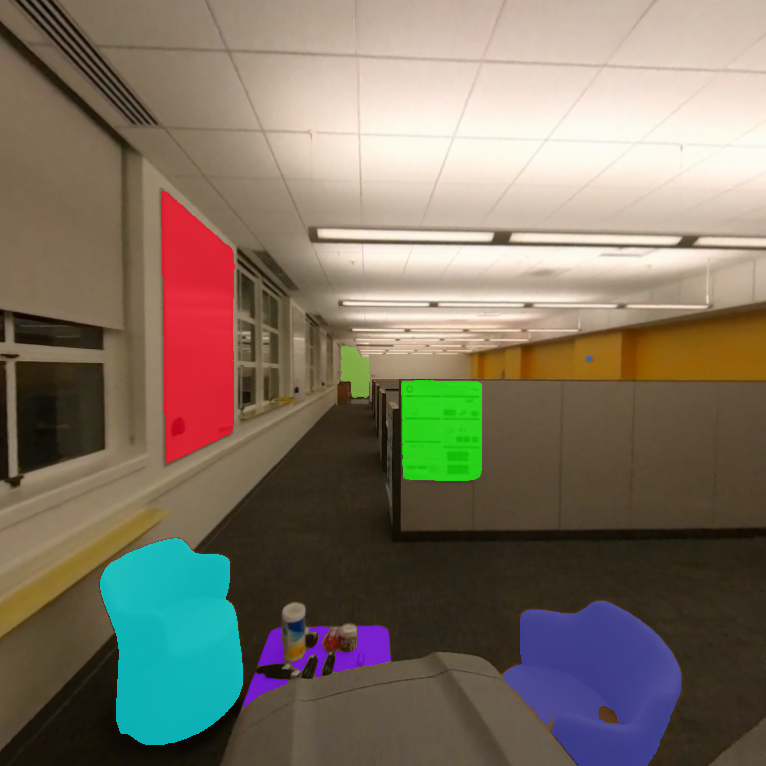}
        \includegraphics[width=0.24\linewidth]{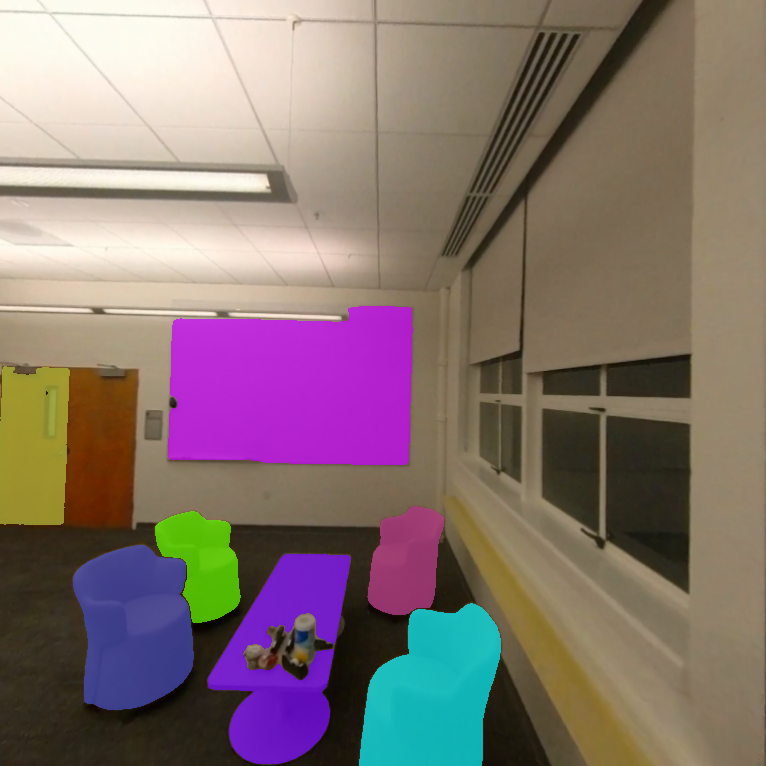}
    \end{minipage}
    \begin{minipage}[c]{0.05\linewidth}
        \centering
        \rotatebox{90}{\footnotesize Ours}
    \end{minipage}%
    \begin{minipage}[c]{0.95\linewidth}
        \centering
        \includegraphics[width=0.24\linewidth]{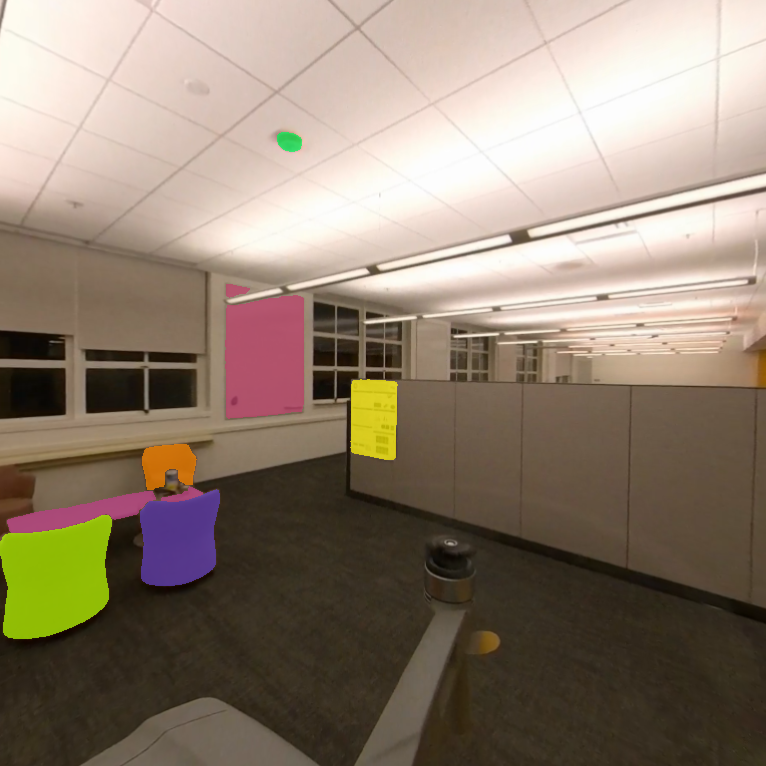}
        \includegraphics[width=0.24\linewidth]{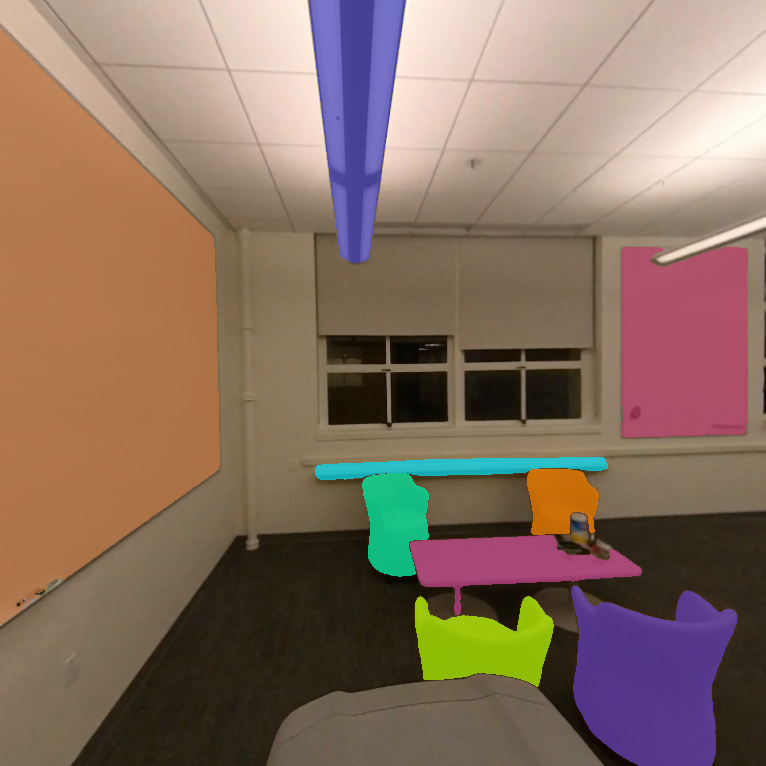}
        \includegraphics[width=0.24\linewidth]{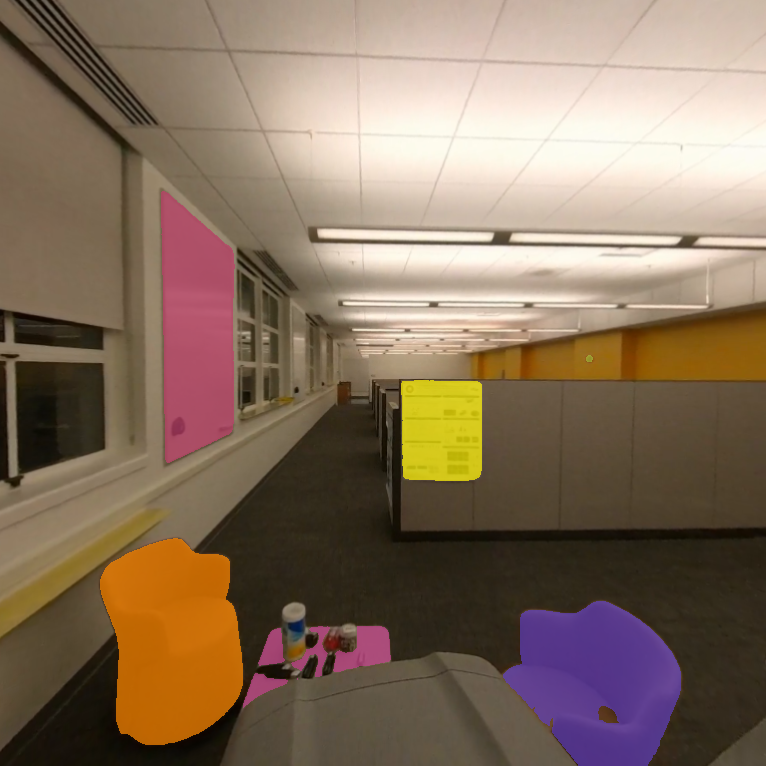}
        \includegraphics[width=0.24\linewidth]{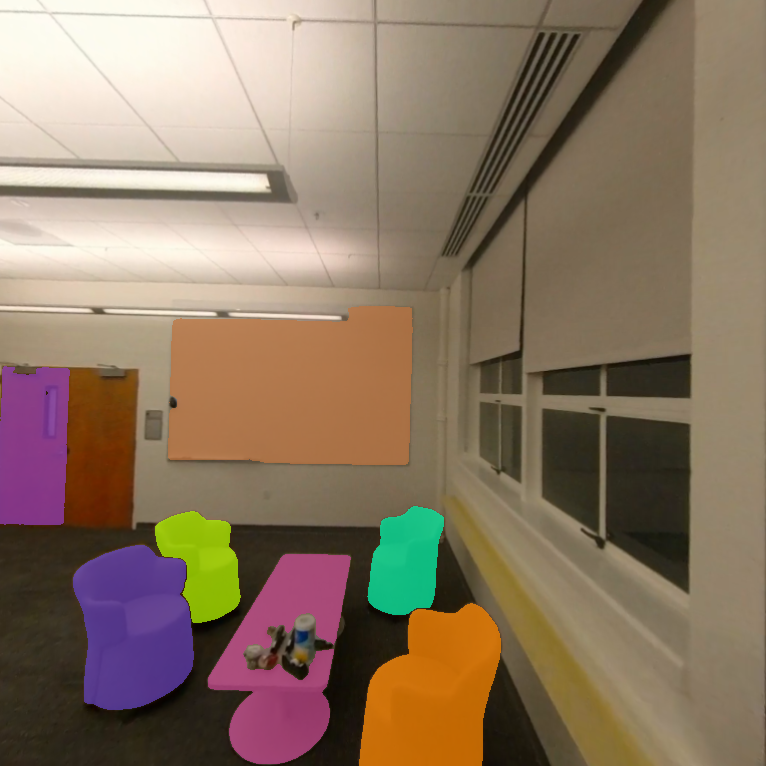}
    \end{minipage}
    \begin{minipage}[c]{0.05\linewidth}
        \centering
        \rotatebox{90}{\footnotesize GAGA \cite{lyu2024gaga}}
    \end{minipage}%
    \begin{minipage}[c]{0.95\linewidth}
        \centering
        \includegraphics[width=0.24\linewidth]{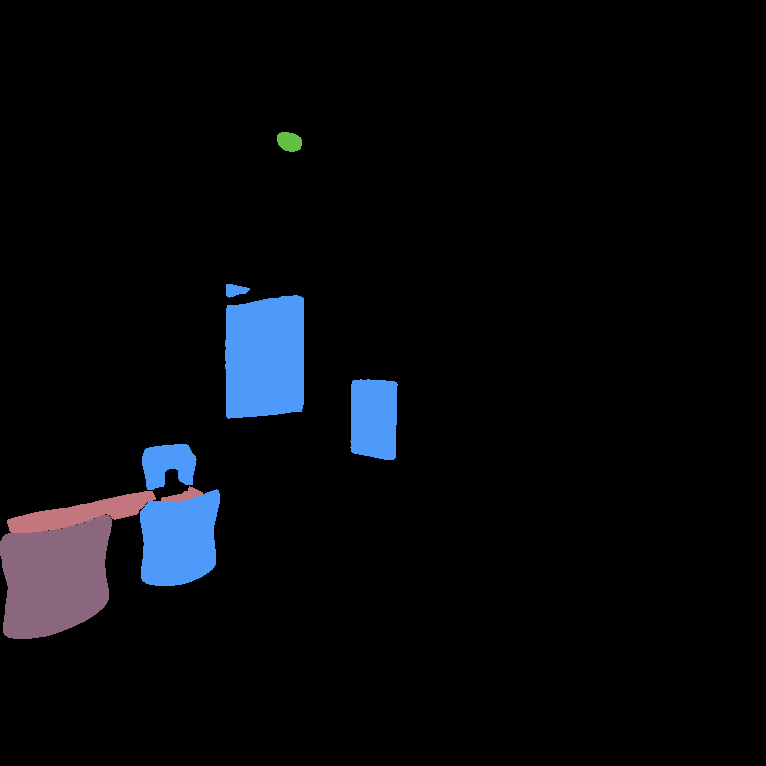}
        \includegraphics[width=0.24\linewidth]{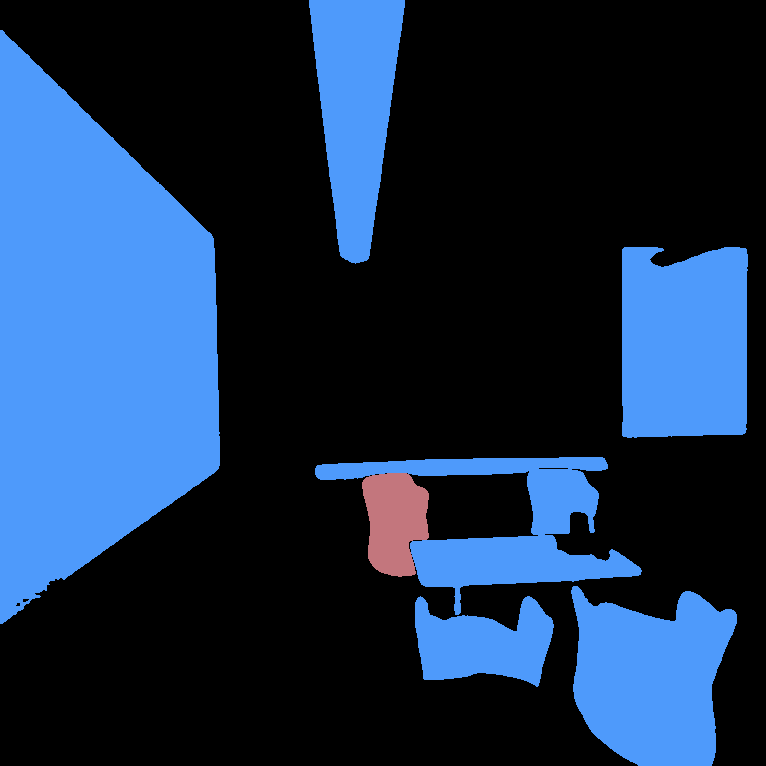}
        \includegraphics[width=0.24\linewidth]{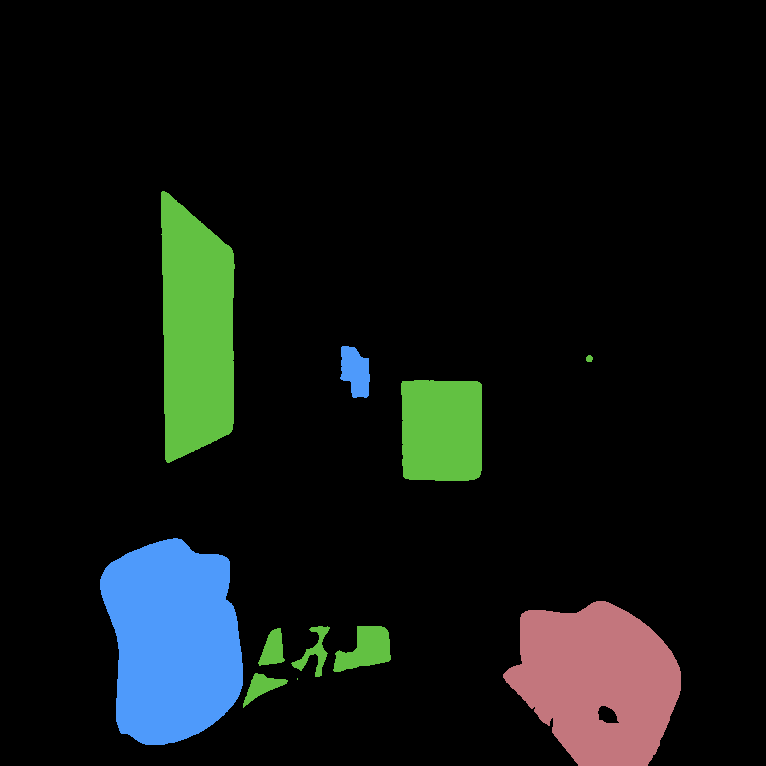}
        \includegraphics[width=0.24\linewidth]{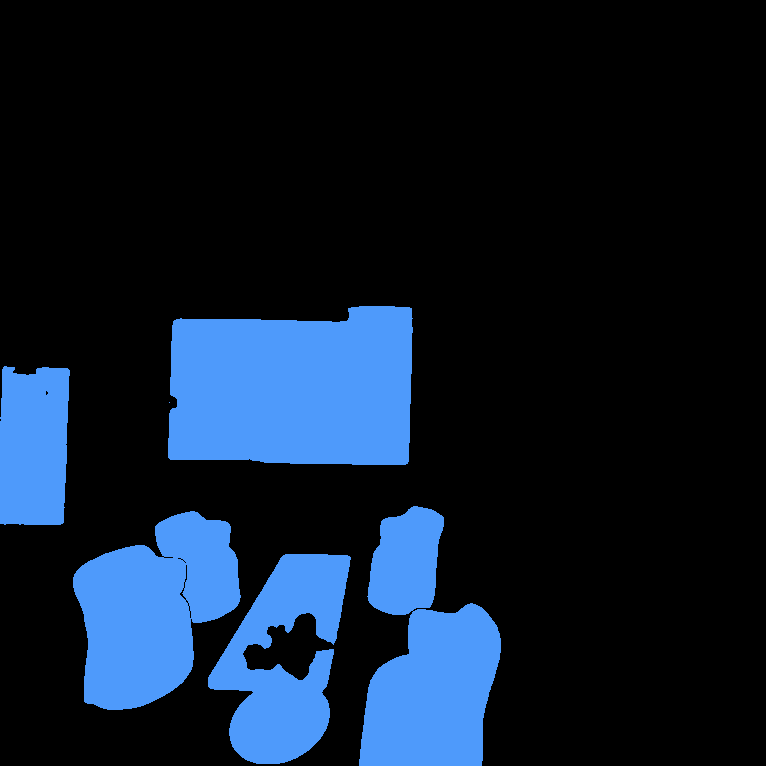}
    \end{minipage}
    \end{minipage}
    \captionof{figure}{We introduce a method that transforms 
    multi-view inconsistent masks, derived from images captured by a drone-mounted $360^\circ$ camera, into a codebook of objects comprised of semantic labels and spatially consistent 3D Gaussian primitives. This produces multi-view consistent masks and outperforms prior methods (right), while also supporting 3D object detection of indoor assets.
    }
    \label{fig:teaser}
    \label{fig:results-maskassoc-cory307}
\end{center}%
}]

\begin{abstract}
We present an approach for object-level detection and segmentation of target indoor assets in 3D Gaussian Splatting (3DGS) scenes, reconstructed from $360^\circ$ drone-captured imagery.
We introduce a 3D object codebook that jointly leverages mask semantics and spatial information of their corresponding Gaussian primitives to guide multi-view mask association and indoor asset detection. 
By integrating 2D object detection and segmentation models with semantically and spatially constrained merging procedures, our method aggregates masks from multiple views into coherent 3D object instances. 
Experiments on two large indoor scenes demonstrate reliable multi-view mask consistency, improving F1 score by 65\% over state-of-the-art baselines, and accurate object-level 3D indoor asset detection, achieving an 11\% mAP gain over baseline methods.
\end{abstract}    
\section{Introduction}
\label{sec:intro}
Accurate 3D reconstruction of indoor environments is valuable across a wide range of applications, such as building mapping and inspection~\cite{xue2021biminspection, xu20253dindoorreconstruction}, augmented and virtual reality~\cite{kang2021tracktagmapaugmentedreality, deng2022foveatednerfvirtualreality, qiu2025advancingxr}, robotics~\cite{byravan2022nerf2realsim2realtransfervisionguided, zhu20243dgsroboticssurvey3dgs}, emergency response planning~\cite{isikdag2013bimemergency}, and cultural heritage preservation~\cite{croce2023culturalheritage, liang2025languagehistoricalbuilding}.

Drone-assisted image capture has recently emerged as an effective means of data capture in indoor environments~\cite{li2023scalable, chen2024scalable}. Owing to its maneuverability and remote operation, drones can capture imagery in cluttered, confined, or hazardous conditions that may be unsafe or impractical for human access~\cite{karam2022dronedisaster}, enabling scalable and repeatable visual data capture.
Recent advances in novel-view synthesis, including Neural Radiance Field (NeRF) representations \cite{mildenhall2020nerf, Barron_2021_mipnerf, Barron_2022_mipnerf360} and 3D Gaussian Splatting (3DGS) \cite{kerbl3Dgaussians}, have enabled high-fidelity 3D scene reconstruction from 2D images. In particular, 3DGS provides an explicit scene representation with real-time rendering capabilities.
Building on this framework, Chen \textit{et al.} \cite{chen2024scalable} proposed a pipeline for reconstructing indoor scenes with 3DGS from $360^\circ$ drone-captured imagery, enabling scalable reconstruction of complex indoor environments.

Beyond geometric reconstruction, semantic understanding of indoor environments is critical for downstream applications such as asset detection~\cite{xu20253dindoorreconstruction}, safety assessment~\cite{liang2026firesafety}, and facility management~\cite{liu2025facilitymanagement, mehraban2025semanticbim}.
Moreover, object-level segmentation provides the necessary information for further downstream tasks such as scene editing and asset isolation.
However, since segmentation masks suffer from inconsistent labeling due to viewpoint changes, achieving multi-view consistent object-level segmentation in large-scale 3D environments remains challenging.
Existing approaches within the 3DGS representation associate 2D masks across views using video tracking models~\cite{ye2024gaussiangrouping, cheng2023deva} or 3D-aware memory banks~\cite{lyu2024gaga}, but these methods are often designed for single-stream sequential data or room-scale scenes and struggle with large indoor spaces containing multiple video streams and reappearing objects.
Furthermore, many prior approaches~\cite{ye2024gaussiangrouping, lyu2024gaga, cen2023saga} adopt a ``segment-everything" paradigm, producing both object and ``stuff" classes such as walls and floors, which is not well-aligned with applications focused on predefined target assets.

In this paper, we present an automated pipeline for detecting and segmenting user-defined indoor assets in large-scale 3DGS scenes.
We introduce an object codebook to associate 2D object masks across views and link them to 3D Gaussian primitives, forming coherent 3D object instances.
To improve robustness, we incorporate confidence-based and spatial filtering to suppress erroneous or noisy masks.
Experimental results show improved multi-view mask label consistency and object-level detection compared to baseline methods, even in dense and occluded indoor scenes.
\section{Related Work}
\label{sec:relatedwork}

\noindent \textbf{3D Gaussian Splatting:}
3D Gaussian Splatting (3DGS)~\cite{kerbl3Dgaussians} represents scenes using learnable anisotropic Gaussian primitives, each associated with parameters that define its geometry and appearance. Because of its explicit point-based formulation and fast inference, 3DGS has become a popular choice for scene reconstruction and novel-view synthesis, and we adopt it as the method of scene representation in this work.

\noindent \textbf{Segment Anything Model:}
The Segment Anything Model (SAM) \cite{kirillov2023segmentanything} is a large-scale foundation model developed for promptable zero-shot image segmentation with strong generalization across domains. While SAM can generate high-quality masks via prompted or automatic segmentation, its outputs are class-agnostic, limiting its use for semantic segmentation without additional components.

\noindent \textbf{Open-Vocabulary 2D Mask Segmentation:}
Grounded SAM \cite{ren2024groundedsam} integrates open-vocabulary detection with SAM-based segmentation by using bounding boxes detected by Grounding DINO \cite{liu2023groundingdino} as prompts for mask generation, producing 2D semantic segmentation masks.

\section{Methodology}
\begin{figure*}[!ht]
    \centering
    \includegraphics[width=0.98\linewidth]{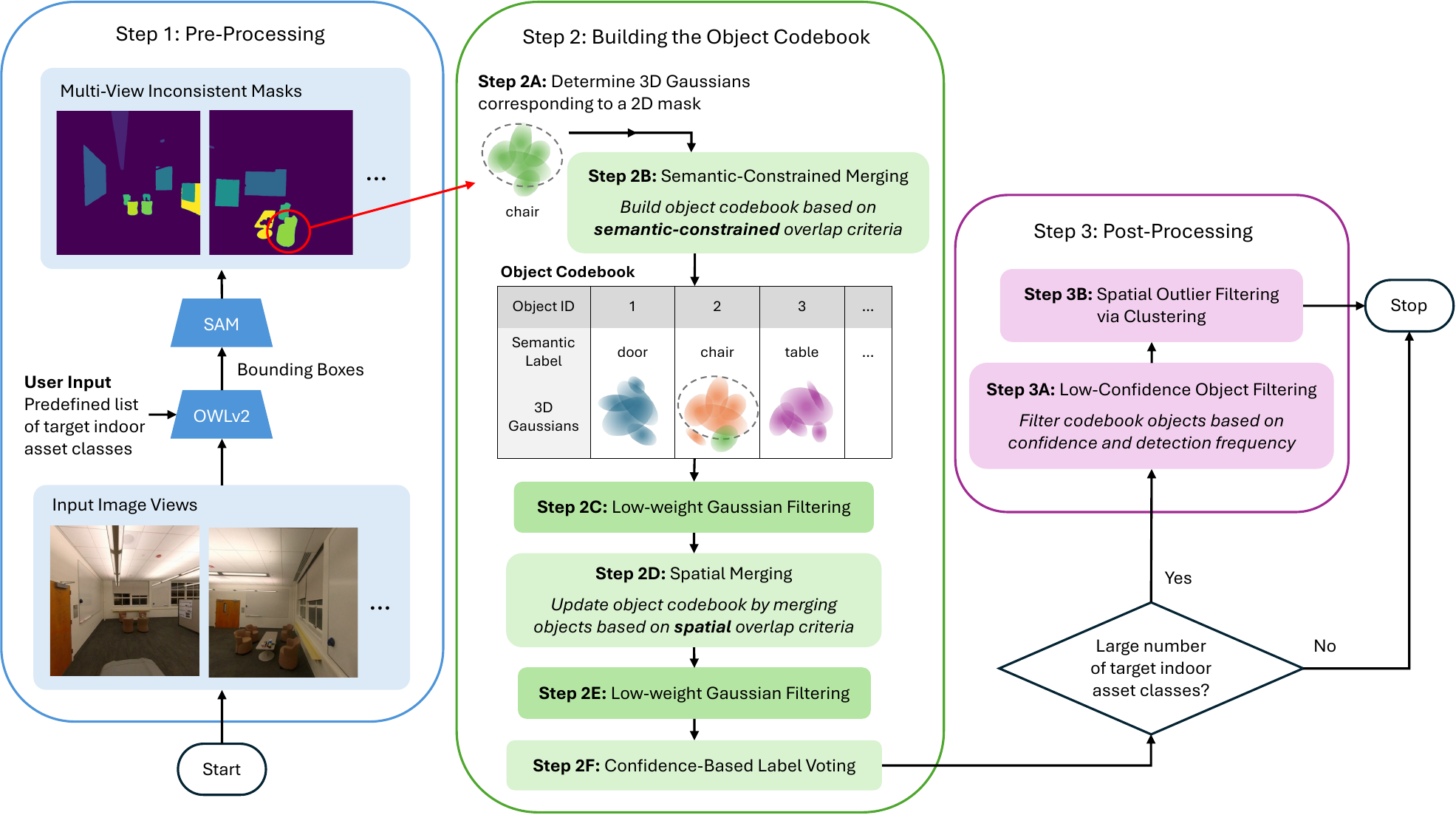}
    \caption{Overview of our proposed pipeline.}
    \label{fig:pipelineoverview}
\end{figure*}

\subsection{Pre-Processing} \label{sec:method-preliminary}
% \paragraph{Labeled Segmentation Masks}
As shown in step 1 of \cref{fig:pipelineoverview}, we pre-process each input view by generating labeled 2D segmentation masks given a predefined list of target object classes.
Following the Grounded SAM paradigm \cite{ren2024groundedsam}, we employ OWLv2 \cite{minderer2023owlv2} for open-vocabulary object detection and classification, followed by SAM \cite{kirillov2023segmentanything} for mask generation.
We replace Grounding DINO \cite{liu2023groundingdino} with OWLv2 due to its improved handling of multi-word class names and its superior performance on fine-grained labels~\cite{bianchi2024devil}. 

Each mask is assigned a nonnegative integer ID, semantic label, and confidence score, defined as the product of its OWLv2 detection and SAM segmentation confidences. To ensure robustness and reliability, we discard masks with low confidence.
Because 2D segmentation masks may still contain missed or incorrect detections and inconsistent IDs across views, we aggregate masks from multiple viewpoints into a unified 3D object codebook. This aggregation improves detection coverage while subsequent filtering and post-processing stages mitigate spurious detections and semantic label inconsistencies.

\subsection{Building the Object Codebook}
\label{sec:method-buildingcodebook}
A recurring challenge of 3DGS segmentation using 2D masks is the lack of mask ID consistency for a given object across different viewpoints.
Ye \textit{et al.}~\cite{ye2024gaussiangrouping} address this issue by passing SAM masks of a sequential image stream to a video tracker~\cite{cheng2023deva}.
Lyu \textit{et al.} on the other hand maintain a bank of objects to spatially associate masks belonging to the same 3D object \cite{lyu2024gaga}.
Inspired by the latter approach, we construct an object codebook, taking advantage of the inherent spatial coherency of 3D objects to associate and aggregate 2D masks across views.

The codebook is a collection of objects, each characterized by a unique object ID, a semantic label, and an associated set of Gaussian primitives. As segmentation masks of individual viewpoints are processed sequentially, the codebook is incrementally populated with newly discovered objects and refined with additional observations. 

\begin{figure}[t]
    \centering
    \begin{subfigure}[t]{0.32\linewidth}
        \centering
        \includegraphics[width=\linewidth]{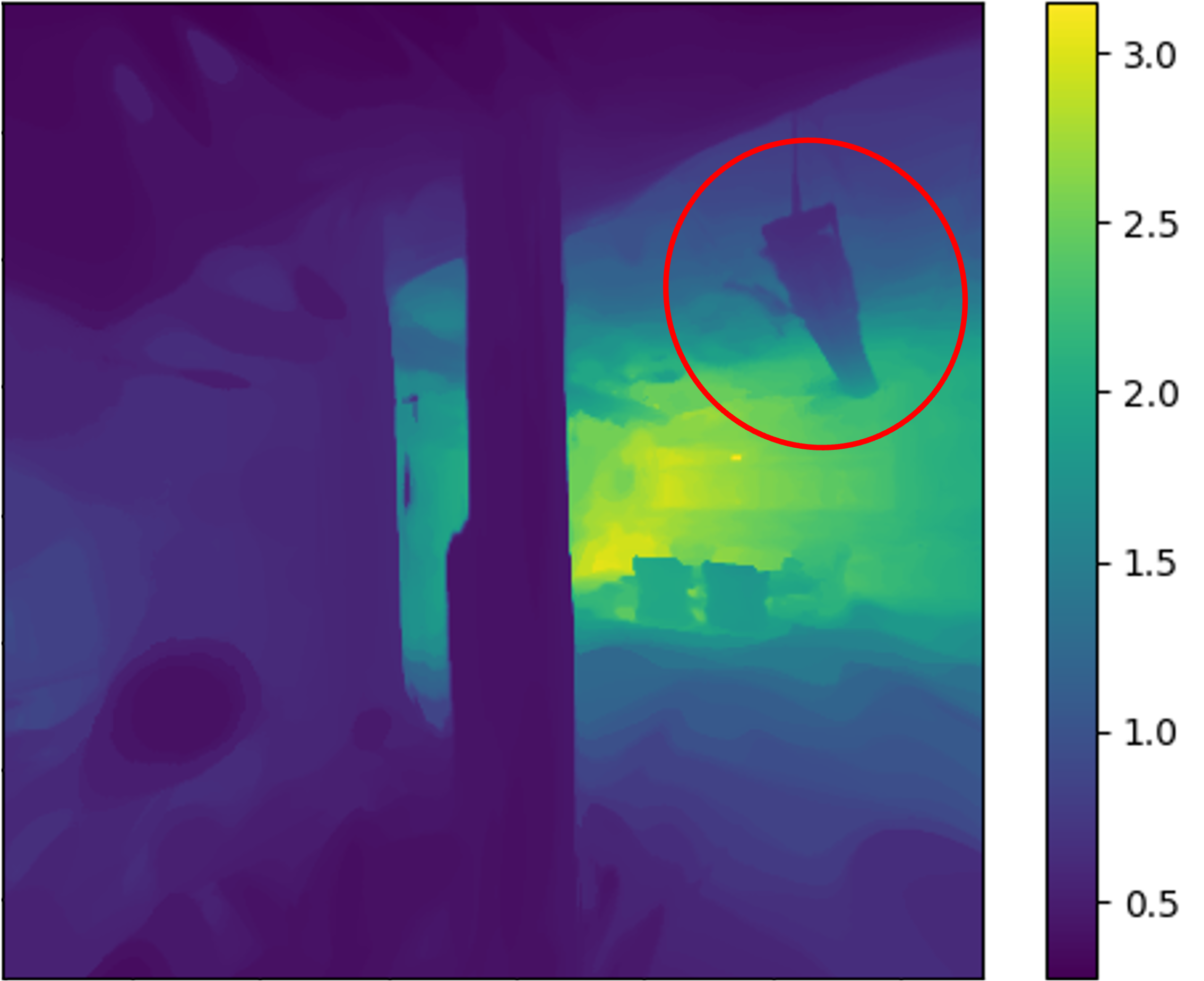}
        \caption{Depth image}
        \label{fig:depthimage}
    \end{subfigure}%
    ~ 
    \begin{subfigure}[t]{0.32\linewidth}
        \centering
        \includegraphics[width=\linewidth]{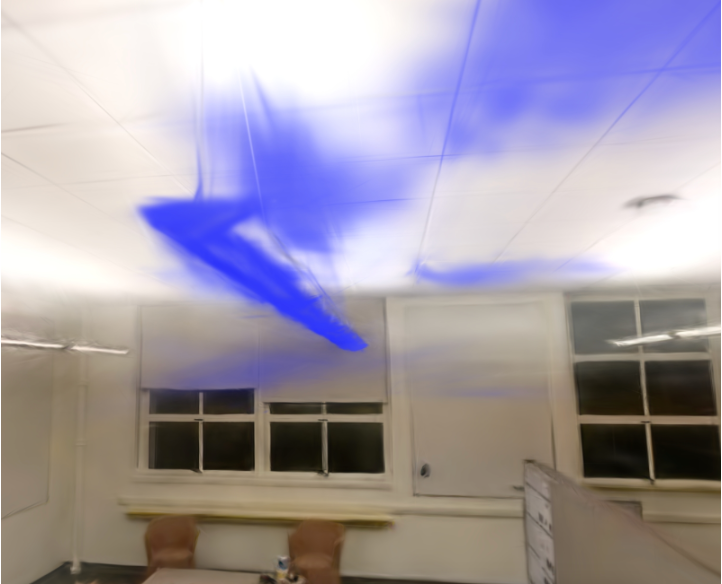}
        \caption{GAGA \cite{lyu2024gaga}}
        \label{fig:depthgaga}
    \end{subfigure}%
    ~ 
    \begin{subfigure}[t]{0.32\linewidth}
        \centering
        \includegraphics[width=\linewidth]{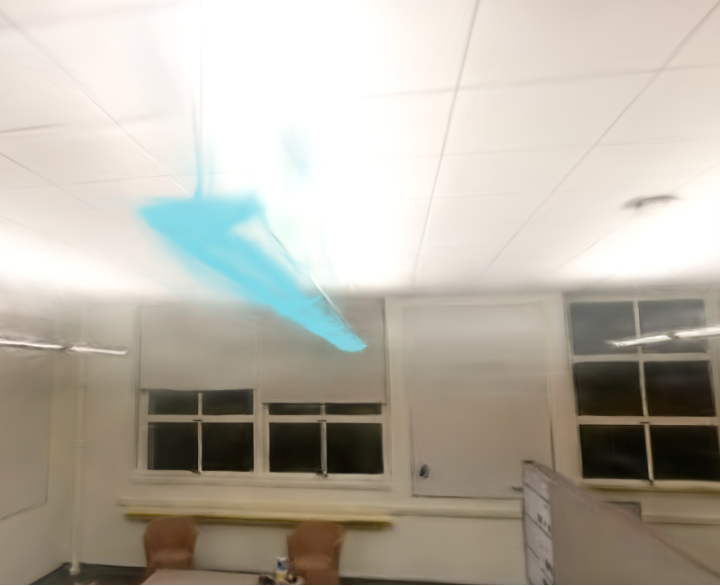}
        \caption{Ours}
        \label{fig:depthours}
    \end{subfigure}
    \caption{Depth-based processing to determine 3D Gaussians corresponding to a 2D mask. (a) Estimated depth image computed via~\cite{luiten2023mediandepth}. The \textit{lamp} object shown in (b) and (c) is circled in red. (b) Results from GAGA's \cite{lyu2024gaga} depth processing method. Gaussians belonging to the ceiling are erroneously included in the \textit{lamp} object's set of Gaussians. (c) Results from our depth processing method.}
    \label{fig:depth}
\end{figure}

\noindent \textbf{3D Gaussians Corresponding to a 2D Mask:}
\label{sec:depthprocessing}
Determining which 3D Gaussians correspond to each 2D masked region, denoted as step 2A in \cref{fig:pipelineoverview}, is a key component of segmenting a 3DGS scene using 2D masks.
Prior work, such as GAGA \cite{lyu2024gaga}, perform this association by selecting Gaussians within a computed depth interval whose centers project onto the masked region.
While effective for masks in scenes with simple compositions, this strategy fails for masks of heavily foreshortened objects that occlude nearby parts of the scene. In such scenarios, the depth interval computed for the mask can be wide enough to erroneously include background Gaussians. As illustrated in the red circled region of \cref{fig:depthimage}, the front section of the ceiling above the lamp possess depth values that fall within the depth interval of the lamp mask computed by GAGA's depth processing scheme. Consequently, the Gaussians on the ceiling are incorrectly included in the lamp's set of Gaussians, shown in \cref{fig:depthgaga}.

To address this limitation, we propose an alternative method for determining Gaussian inliers that adapts to the local depth-variation within a masked region, preventing the inclusion of background Gaussians, as depicted in \cref{fig:depthours}.
We first render a depth image $D$ following the approach of Luiten \textit{et al.} \cite{luiten2023mediandepth, luiten2023dynamic}, shown in \cref{fig:depthimage}.
For each Gaussian whose center $\mu$ projects to a 2D point $(x_p, y_p)$ within the masked region $m$, we compare $\mu$'s depth from the camera, denoted as $d_{\mu}$, with the depth-image value at the projected point, $D(x_p, y_p)$.
A Gaussian is marked as an inlier if its depth $d_{\mu}$ and its depth-image value $D(x_p, y_p)$ are within an adaptive tolerance value $\delta(x_p, y_p)$,
\begin{equation}
    d_\mu \leq D(x_p, y_p) \pm \delta(x_p, y_p).
\end{equation}
The tolerance $\delta$ at each pixel position $(x, y)$ represents the local depth variation in the neighborhood of $(x, y)$. We define this neighborhood as the pixels that are both within a $7\times7$ window centered at the position $(x,y)$ and lie within the masked region $m$:
\begin{equation}
\footnotesize
\mathcal{N}_{(x, y)} :=
\left\{
  (n_x, n_y) \;\middle|\;
  \begin{gathered}
    n_x \in \{x-3, \dots, x+3 \}, \; n_x \neq x, \\
    n_y \in \{y-3, \dots, y+3 \}, \; n_y \neq y, \\
    (n_x, n_y) \in m
  \end{gathered}
\right\}.
\end{equation}

To compute the adaptive tolerance $\delta$, for every pixel position $(x, y)$ in the masked region of the depth image $D$, we evaluate the maximum absolute difference between the depth at $(x, y)$ and its neighbors $\mathcal{N}_{(x, y)}$,
\begin{equation}
\footnotesize
    \delta(x, y) = \max \{ |D(x, y) - D(n_x, n_y)|
    \mid (n_x, n_y) \in \mathcal{N}_{(x, y)}\}
\end{equation}
This formulation provides a per-pixel estimate of the local depth range, allowing the tolerance to adapt to the surface geometry of the masked object. Regions with low depth variation yield tighter tolerances, whereas regions with larger depth variations yield larger, but bounded, tolerances. To prevent excessively permissive tolerances in regions with large depth disparities, we impose a fixed upper bound on the tolerance; this avoids imprecise or unstable masks near occlusion boundaries from over-including distant Gaussians.
The resulting adaptive tolerance at a pixel coordinate $(x, y)$ is defined as
\begin{equation}
% \begin{aligned}
\footnotesize
\delta(x, y) := \max \Bigl\{
    \lvert D(x, y) - D(n_x, n_y) \rvert \Bigm|
    \substack{
        (n_x, n_y) \in \mathcal{N}_{(x, y)} \\
        \lvert D(x, y) - D(n_x, n_y) \rvert \leq T
    }
\Bigr\}
% \end{aligned}
\end{equation}
where $T$ is empirically set to 0.5 as an upper bound on the depth tolerance. In practice, most depth disparities remain below this threshold. Cases where this upper bound is exceeded are attributable to noisy masks that adversely affect depth estimates; these disparities are therefore intentionally excluded.

\noindent \textbf{Semantic-Constrained Merging:} \label{sec:semanticmerging}
In this section, corresponding to step 2B of \cref{fig:pipelineoverview}, we describe the procedure for processing 2D segmentation masks to incrementally populate and update the object codebook.

For a given image mask $m$, we compare its corresponding 3D Gaussians, denoted by $\mathcal{G}(m)$, against each existing object in the codebook that shares a semantic label. Let $\mathcal{G}_i$ denote the set of Gaussians associated with object $i$.
If the overlap between $\mathcal{G}(m)$ and $\mathcal{G}_i$ exceeds a predefined threshold $\tau_{overlap}$, selected through the ablation study described in \cref{sec:ablationthresholds}, then the Gaussians $\mathcal{G}(m)$ are merged into object~$i$:
$\mathcal{G}_i \leftarrow \mathcal{G}_i \cup \mathcal{G}(m).$
Otherwise, a new object is created in the codebook and initialized with the mask's Gaussians $\mathcal{G}(m)$ and corresponding semantic label.
We adopt the Gaussian overlap metric proposed in GAGA \cite{lyu2024gaga}.

We require that a mask and an existing codebook object share the same semantic label before computing the overlap between their associated Gaussians. Conditioning on semantic label mitigates erroneous merges by preventing the fusion of spatially proximate but semantically-distinct objects.
More importantly, semantic conditioning is vital in preventing masks of smaller objects from being absorbed into existing larger objects.
For instance, consider a \textit{door} object already stored in the object codebook and a mask corresponding to a \textit{window} belonging to that door. Without semantic conditioning, the \textit{window}'s Gaussians, likely a subset of the \textit{door}'s Gaussians, would yield an overlap ratio close to 1.0 and incorrectly trigger a merge. Enforcing semantic consistency prevents such unintended merges.

\noindent \textbf{Low-Weight Gaussian Filtering:}
\label{sec:gaussianfiltering}
After constructing the object codebook, we refine each object's Gaussians to reduce the impact of imprecise segmentation masks, particularly of distant objects.
As shown in \cref{fig:pipelineoverview}, this filtering is applied in two stages: step~2C and step~2E.

Each mask possesses a confidence score, assigned during mask generation.
Since masks corresponding to distant objects are generally less reliable, we further penalize them.
The weight of a mask is defined as its confidence score divided by its estimated depth, computed as the average of the mask's depth image values.

For each Gaussian $G$, its weight $w_G$ is accumulated from the weights of the masks the Gaussian is associated with.
For a given object, Gaussians with low accumulated weight correspond to one or more of the following cases: they originate from distant imprecise masks, are supported by only a few observations, or are derived from masks with low confidence. Such Gaussians are more likely to be spurious. We prune these unreliable points from an object's set of Gaussians by first determining the maximum Gaussian weight within the object, $w_{\max}$.
Then, any Gaussian $G$ whose weight falls below a relative threshold $\tau_{filter} \in (0, 1)$ is removed: 
$w_G < w_{\max} \cdot \tau_{filter}$,
where $\tau_{filter}$ is determined via ablation study described in \cref{sec:ablationthresholds}.

This relative threshold filtering strategy adapts to variations of observation frequency and confidence across different objects, allowing high-confidence and more frequently observed Gaussians to be retained while pruning Gaussians that arise from noisy, distant, or weakly supported masks.

\noindent \textbf{Spatial Merging of Objects:}
\label{sec:spatialmerging}
After semantic-conditioned merging, all segmentation masks have been incorporated into the object codebook.
However, due to the strict label-matching constraint, semantic label inconsistencies across views from mask generation can cause the same 3D object to be represented as multiple distinct codebook entries.
The Spatial Merging stage, denoted as step~2D in \cref{fig:pipelineoverview}, addresses this issue by identifying and merging such duplicate objects based solely on the spatial overlap of their associated Gaussians.

During semantic-conditioned merging, partial 2D observations are aggregated to form increasingly complete 3D object geometry, enabling the subsequent spatial merging stage to apply stricter overlap criteria.
To this end, we construct an undirected graph where each object in the codebook is represented by a vertex. An edge is formed between two vertices $A$ and $B$ if their Gaussians, $\mathcal{G}_A$ and $\mathcal{G}_B$, exhibit sufficient overlap, as defined by the following symmetric conditions,
\begin{equation}
\footnotesize
% \begin{split}
    % &\text{Overlap}(A, B) = 
    \frac{|\mathcal{G}_A \cap \mathcal{G}_B|}{|\mathcal{G}_A|} > \tau_{spatial}
    \quad\text{and}\quad 
    % &\text{Overlap}(B, A) = 
    \frac{|\mathcal{G}_B \cap \mathcal{G}_A|}{|\mathcal{G}_B|} > \tau_{spatial},
% \end{split}
\end{equation}
where $\tau_{spatial}$ is a predefined merging threshold, determined through ablation study described in \cref{sec:ablationthresholds}. Requiring the overlap criteria to hold in both directions ensures that neither object is largely subsumed by the other. 

After constructing the graph, we identify all of its connected components. Each connected component $\mathcal{C}$ consists of vertices $v$ representing the objects whose associated Gaussians sufficiently overlap and should thus be merged. Merging the objects within a connected component produces a single object whose associated Gaussian set is given by the union of the Gaussian sets of all objects in that component.
As a result, the codebook is transformed into a collection of merged objects, with one object for each connected component of the graph. Each merged object now consists of a set of Gaussians and a list of semantic labels inherited from the segmentation masks that contributed to the object's construction.

\noindent \textbf{Confidence-Based Label Voting:}
\label{sec:labelvoting}
After merging objects spatially, each object is associated with a list of masks along with their corresponding semantic labels and confidence scores. 
Since this list of masks may contain inconsistent semantic labels, we determine a final object-level semantic label using a confidence-based voting scheme, denoted as step 2F in \cref{fig:pipelineoverview}.
To do so, we compute a total confidence score sum for each distinct class label and choose the label with the highest total score.
This voting strategy favors labels that are consistently supported by high-confidence masks and limits the influence of isolated or low-confidence misclassifications. 

\subsection{Post-Processing}
\label{sec:method-postprocessing}
For complex scenes with a large number of object classes, additional post-processing is required to improve object detection and segmentation reliability.
Empirically, increasing the number of candidate object classes for the 2D object detector raises the likelihood of false positive detections and semantic mislabels. Therefore, we apply the following steps, altogether denoted as step 3 in \cref{fig:pipelineoverview}, to scenes where we detect a large number of class labels, e.g. more than 10 class labels.

\noindent \textbf{Low-Confidence Object Filtering:}
\label{sec:objfiltering}
Erroneous segmentation masks may give rise to spurious objects in the object codebook.
To remove such objects, we employ an object-level confidence metric to filter weakly supported objects, shown as step 3A in \cref{fig:pipelineoverview}.
We define an object-level confidence score that accounts for both detection confidence and frequency across views.
Since true objects are typically observed from multiple viewpoints, which is particularly true for our $360^\circ$ drone-captured data, we compute the confidence of an object $O$ derived from the set of masks $\mathcal{M}$ as
\begin{equation}
    \footnotesize
    c_O = \log(|\mathcal{M}|) \cdot \frac{1}{|\mathcal{M}|}\sum_{m \in \mathcal{M}} c_m
\end{equation}
where $c_m$ is the confidence score of a mask $m$. 
Objects whose confidence fall below a threshold $\tau_{object}$ are removed from the codebook, with $\tau_{object}$ determined via ablation described in \cref{sec:ablationthresholds}.
This step suppresses spurious objects while preserving objects that are consistently and confidently detected.

\noindent \textbf{Spatial Outlier Filtering via Clustering:}
While earlier stages filtered Gaussians on a per-object basis using confidence-based weighting, dense scenes may still contain spatial outliers.
As shown in \cref{fig:pipelineoverview}'s step~3B, we apply HDBSCAN($\hat{\epsilon}$) clustering \cite{malzer2020cluster} to each object's set of Gaussian centers.
The distance threshold $\epsilon$ is estimated per object using a \textit{sorted k-dist graph}, following DBSCAN~\cite{ester1996dbscan}. For 3D data, we set $minPts=6$ \cite{sander1998gdbscan}, and select $\epsilon$ at the elbow of the k-dist curve, detected automatically using the $\texttt{kneed}$ package \cite{arvkevi2018kneed} which implements the Kneedle algorithm~\cite{satopaa2011kneedle}.
Points classified as outliers by HDBSCAN($\hat{\epsilon})$ or are assigned low cluster membership probability are removed, yielding more spatially consistent object geometry.

\section{Experimental Results}

\subsection{Datasets} \label{sec:datasets}
\begin{figure}[!t]
    \centering
    \begin{subfigure}[t]{0.35\linewidth}
        \centering
        \includegraphics[width=\linewidth]{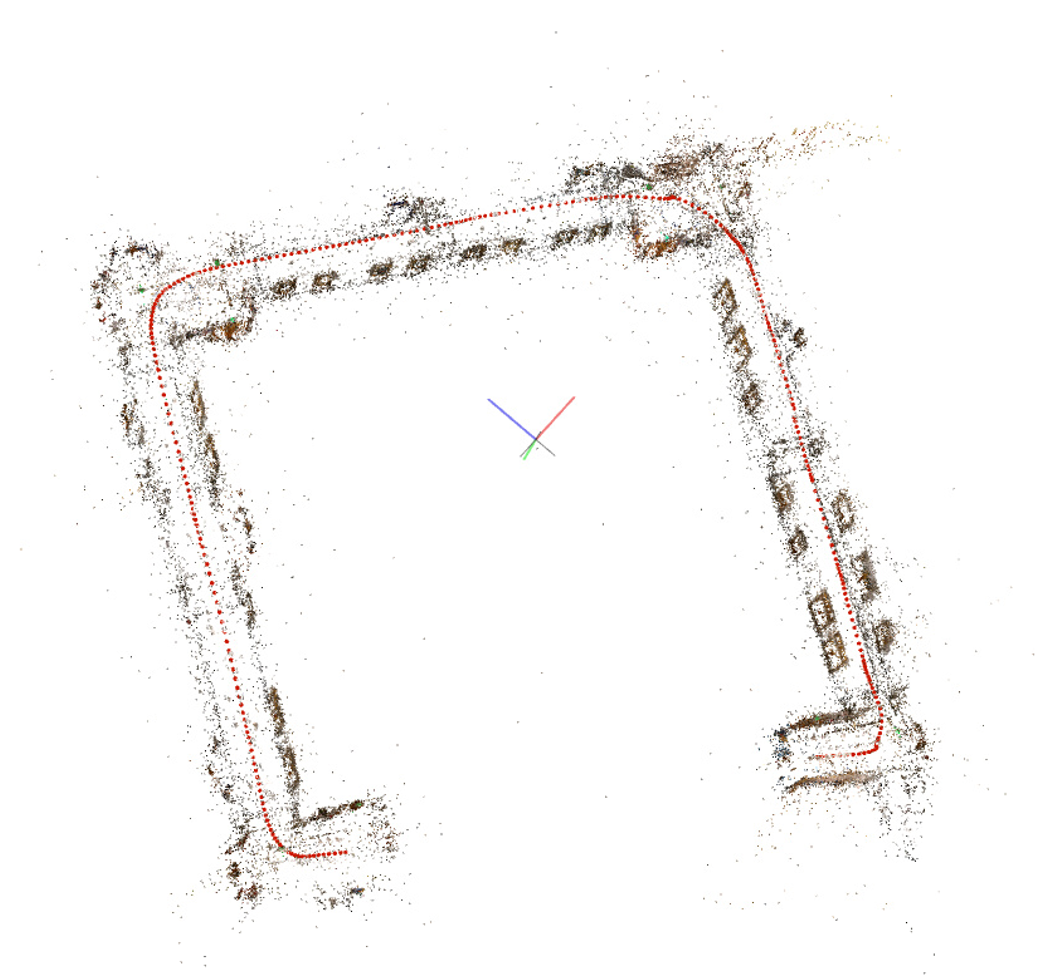}
        \caption{Cory 3rd Floor}
        \label{fig:colmap_cory3rdfloor}
    \end{subfigure}%
    \begin{subfigure}[t]{0.65\linewidth}
        \centering
        \includegraphics[width=\linewidth]{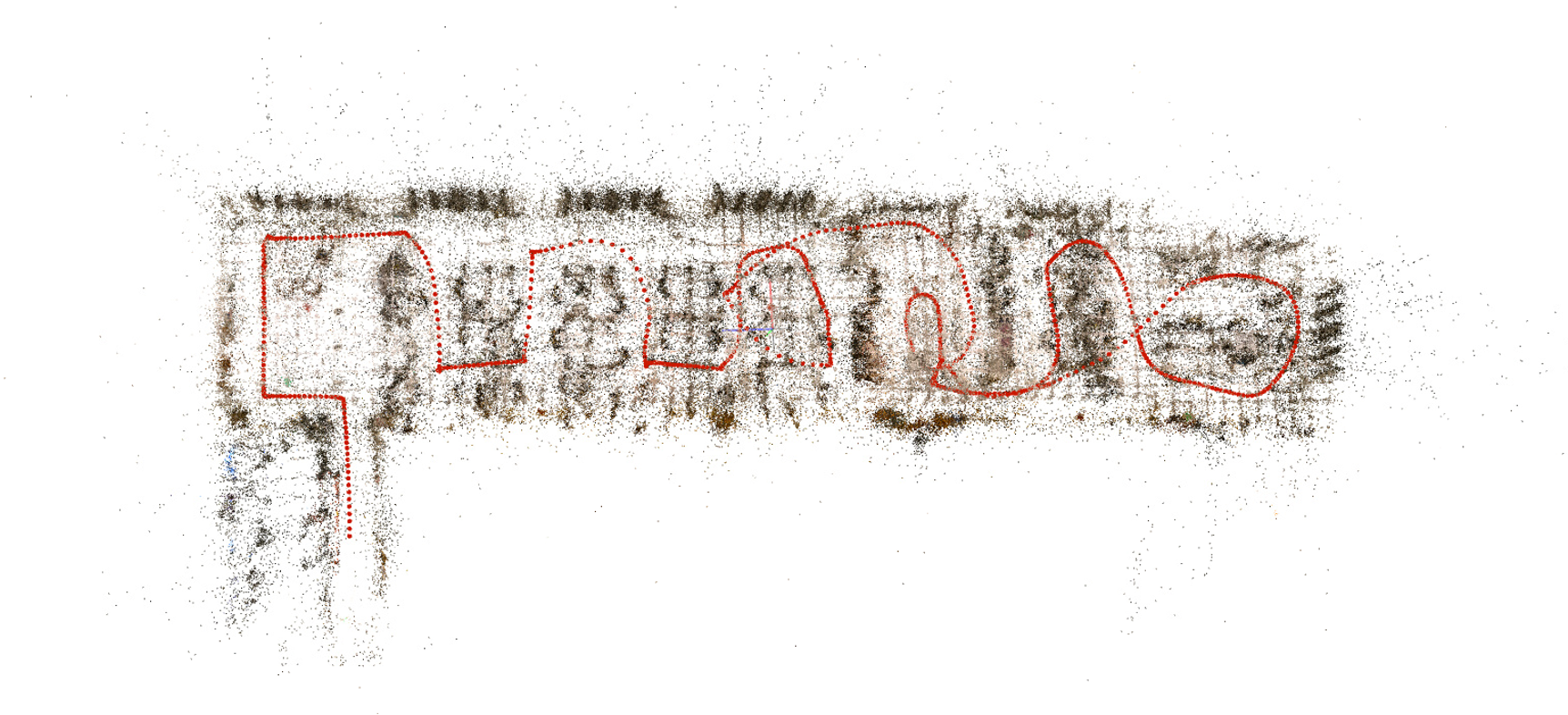}
        \caption{Cory 307 Office}
        \label{fig:colmap_cory307office}
    \end{subfigure}
    \caption{Sparse point clouds reconstructed by COLMAP \cite{schoenberger2016colmap} for both datasets. Camera positions are displayed in red, tracking the drone trajectory during data capture.}
    \label{fig:colmap}
\end{figure}

We evaluate the proposed approach on two large-scale indoor datasets captured in Cory Hall, an academic building at University of California, Berkeley.
Indoor Scene 1 (Cory 3rd Floor) consists of 2232 input views capturing three interconnected corridors.
Indoor Scene 2 (Cory 307 Office) contains 6532 input views of Room 307, an office environment with two open-plan areas connected by a shared breakroom.
\Cref{fig:colmap} shows the corresponding sparse point clouds and camera poses reconstructed using COLMAP~\cite{schoenberger2016colmap}.

\noindent \textbf{Data Acquisition and Processing:} \label{dataacquisitionprocessing}
Both datasets were captured using a $360^\circ$ video camera mounted on a drone flown by a human pilot.
Following Chen \textit{et al.}~\cite{chen2024scalable}, raw spherical frames are projected into eight cube-face images per timestep.
Since the drone body can appear in the projected cube-face views, we adopt Chen \textit{et al.}'s inpainting strategy for the Cory 3rd Floor dataset, where the drone occupies feature-sparse regions.
For the denser Cory 307 Office dataset, inpainting may introduce artifacts; we instead segment the drone using SAM~\cite{kirillov2023segmentanything} to exclude its pixels from 3DGS optimization.

\noindent \textbf{Indoor Asset Labels:}
For each dataset, we provide OWLv2~\cite{minderer2023owlv2} with a predefined list of target indoor asset classes in the form of a text prompt. The class labels used for each dataset are listed in \cref{tab:classlabels}.

\begin{table}[!h]
    \centering
    \caption{Indoor asset class labels. A total of 8 classes are used for the Cory 3rd Floor dataset and 29 classes for Cory 307 Office.}
    \label{tab:classlabels}
    \footnotesize
    \setlength\tabcolsep{4pt}
    \begin{tabular}{@{}l l l l@{}}
        \toprule
        \multicolumn{4}{c}{Cory 3rd Floor Class Labels} \\
        \midrule
        door&drinking fountain&exit sign&fire alarm \\
        fire extinguisher&lamp&television&window \\
        \bottomrule
    \addlinespace[0.5em]
        % \toprule
        \multicolumn{4}{c}{Cory 307 Office Class Labels} \\
        \midrule
        bench& bottle&chair&computer keyboard \\
        computer monitor&computer mouse&cupboard&desk \\
        door&exit sign&file cabinet&fire alarm \\
        fire extinguisher&headphones&ladder&lamp \\
        laptop&microwave&mug&oven \\
        poster&printer&refrigerator&sink \\
        table&telephone&trash can&whiteboard \\
        window&&& \\
        \bottomrule
    \end{tabular}
\end{table}
%======================================================================
\subsection{Implementation Details} \label{implementationdetails}
We acquire spherical $360^\circ$ video data at 30 FPS for both datasets using an Insta360 ONE RS camera mounted on a DJI Mavic Air 2 drone.
Insta360 Studio is then used to convert spherical $360^\circ$ imagery into $5760 \times 2800$ equirectangular MP4 video before its frames are extracted at 3 FPS and projected onto cube faces, forming $768 \times 768$ resolution images.
We use COLMAP \cite{schoenberger2016colmap} as the SfM method.
The 3D Gaussian Splatting models for all experiments are trained for 30K iterations using vanilla 3DGS \cite{kerbl3Dgaussians}.
For 2D object detection, we use the weight-space ensemble of the self-trained and fine-tuned OWLv2 checkpoints~\cite{minderer2023owlv2} that employ the CLIP~L/14~\cite{radford2021clip} backbone.
For image segmentation, we use SAM~\cite{kirillov2023segmentanything} with the ViT-H backbone~\cite{li2022exploringplainvisiontransformer}.
Thresholds used for each pipeline step are shown in \cref{tab:thresholds}.
All experiments, implemented in PyTorch~\cite{Ansel_PyTorch}, are performed on a single 24GB NVIDIA TITAN RTX GPU.

\begin{table}[t]
    \centering
    \caption{Thresholds used for each pipeline step from \cref{fig:pipelineoverview}}
    \label{tab:thresholds}
    \footnotesize
    \begin{tabular}{@{}l l c@{}}
        \toprule
         Pipeline Step & Symbol & Threshold \\
         \midrule
         2B \; Semantic-Constrained Overlap & $\tau_{overlap}$ & 0.2 \\
         2C \; $1^\text{st}$ Low-Weight Gaussian Filtering & $\tau_{filter1}$ & 0.4 \\
         2D \; Spatial Merging Overlap & $\tau_{spatial}$ & 0.3 \\
         2E \; $2^\text{nd}$ Low-Weight Gaussian Filtering & $\tau_{filter2}$ & 0.3 \\
         3A \; Object Filtering & $\tau_{object}$ & 0.8 \\
         \bottomrule
    \end{tabular}
\end{table}
%======================================================================

\subsection{Evaluation Setup} \label{evalmetricssetup}
\noindent \textbf{Evaluation Study 1 - 2D Mask Association:}
We first evaluate our multi-view mask association strategy.
Using the final object codebook (after the \textit{Stop} marker in \cref{fig:pipelineoverview}), we relabel each 2D mask with the object ID of its associated codebook entry, producing object-consistent masks.
We report mIoU, Precision, Recall, and F1-score at an IoU threshold of 0.5, along with pipeline runtime required to obtain the multi-view consistent masks.

\noindent \textbf{Evaluation Study 2 - Object Detection:}
We assess object detection performance using mean Average Precision (mAP) and mean Log Average Miss Rate (mLAMR). For each test viewpoint, we render 3D objects to generate labeled 2D bounding boxes by enclosing their projected extents and assigning their corresponding semantic labels.

\noindent \textbf{Ground-Truth Annotation:}
We generate ground-truth annotations for both tasks, reserving $10\%$ of each dataset for testing ($5\%$ for object detection on Cory 307 Office due to high cost of annotation).
For mask association, we manually relabel the segmentation masks produced in \cref{sec:method-preliminary} to enforce consistent object IDs across views.
For object detection, we annotate test images with 2D bounding boxes for the target classes listed in \cref{tab:classlabels} using the CVAT annotation tool~\cite{2023cvat}.

\subsection{Results} \label{results}
\begin{table}[t]
\centering
\footnotesize
\caption{Evaluation results for 2D mask association.}
\label{tab:results_maskassociation}
\begin{tabular}{ l *5{S[table-format=2.2]} }
\toprule
& \multicolumn{5}{c}{Cory 3rd Floor} \\
\cmidrule(lr){2-6}
Method & {mIoU} & {Precision} & {Recall} & {F1} & {Runtime $\downarrow$} \\
\midrule
Ours & \textbf{75.84} & \textbf{82.61} & \textbf{84.07} & \textbf{83.33} & \textbf{10min}\\
GAGA \cite{lyu2024gaga} & 9.28 & 73.33 & 9.73 & 17.19 & \text{39min}\\
\toprule
& \multicolumn{5}{c}{Cory 307 Office} \\
\cmidrule(lr){2-6}
Method & {mIoU} & {Precision} & {Recall} & {F1} & {Runtime $\downarrow$} \\
\midrule
Ours & \textbf{66.01} & \textbf{67.05} & \textbf{72.54} & \textbf{69.69} & \textbf{1hr 33min}\\
GAGA \cite{lyu2024gaga} & 2.10 & 33.33 & 1.64 & 3.12 & {3hr 43min}\\
\bottomrule
\end{tabular}
\end{table}
%========================================
\begin{table}[t]
\centering
\caption{Number of unique masks that appear in the 2D mask-association test sets}
\label{tab:results_numberofmasks}
\footnotesize
\begin{tabular}{ l *2{S[table-format=3.0]} }
\toprule
& \multicolumn{1}{c}{Cory 3rd Floor} & \multicolumn{1}{c}{Cory 307 Office} \\
\midrule
Ground Truth & 113 & 244 \\
Ours & 115 & 264 \\
GAGA \cite{lyu2024gaga} & 15 & 12 \\
\bottomrule
\end{tabular}
\end{table}
%========== RESULT FOR MASK ASSOCIATION =========
% Full Masks -- Cory 3rd Floor (GAGA)
\begin{figure}
    \centering
    \begin{subfigure}{0.45\linewidth}
        \centering
        \includegraphics[width=0.7\linewidth]{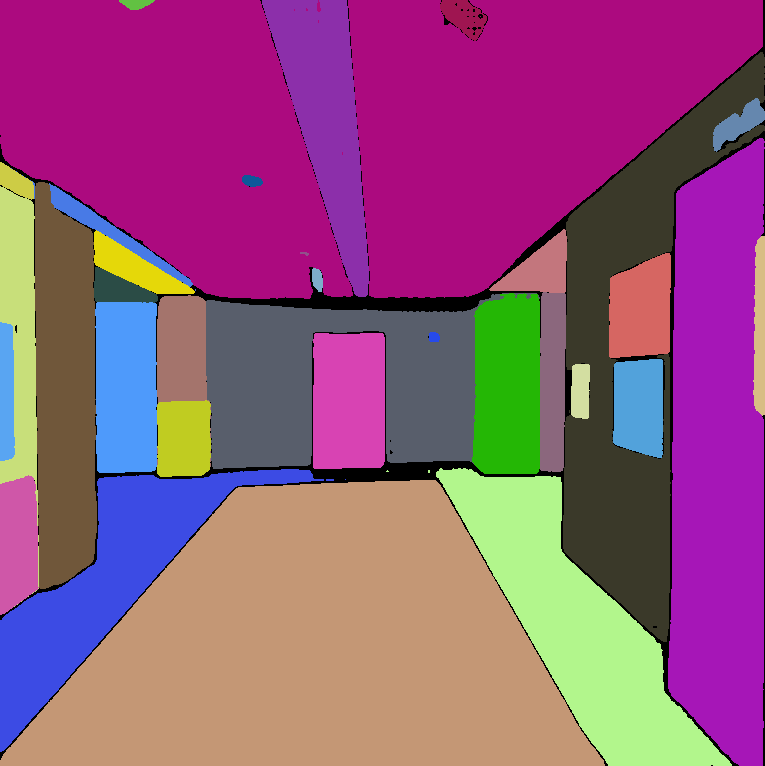}
        \caption{Beginning of Image Sequence}
    \end{subfigure} %
    \begin{subfigure}{0.45\linewidth}
        \centering
        \includegraphics[width=0.7\linewidth]{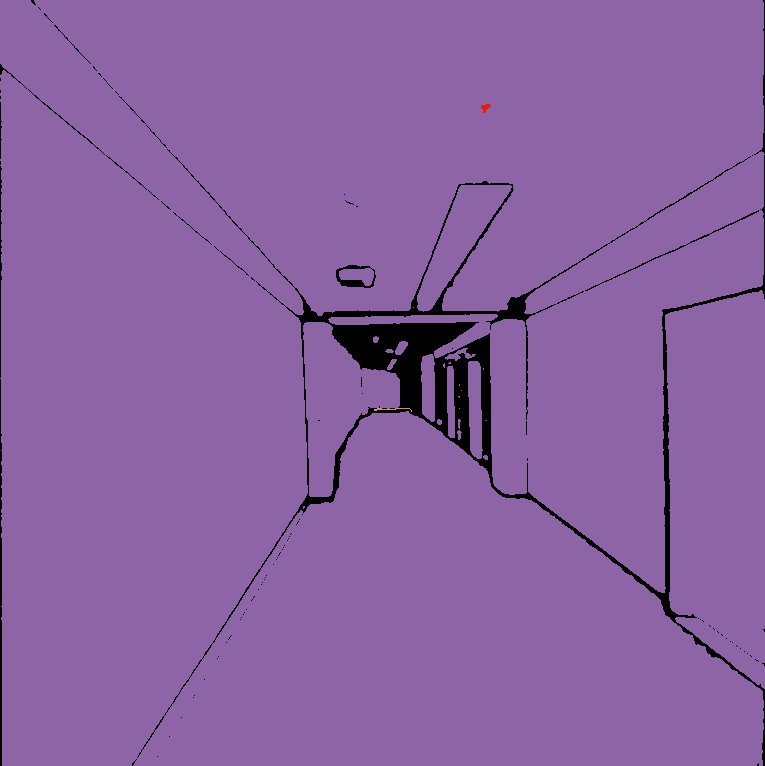}
        \caption{Later in Image Sequence}
    \end{subfigure} %
    \caption{Mask association results produced by GAGA \cite{lyu2024gaga} using inputs from SAM's \textit{segment-everything} mode.}
    \label{fig:results-gaga-fullseg}
\end{figure}
% Temporal plots
\begin{figure}
    \centering
    \begin{subfigure}{\linewidth}
        \centering
        \includegraphics[width=0.9\linewidth]{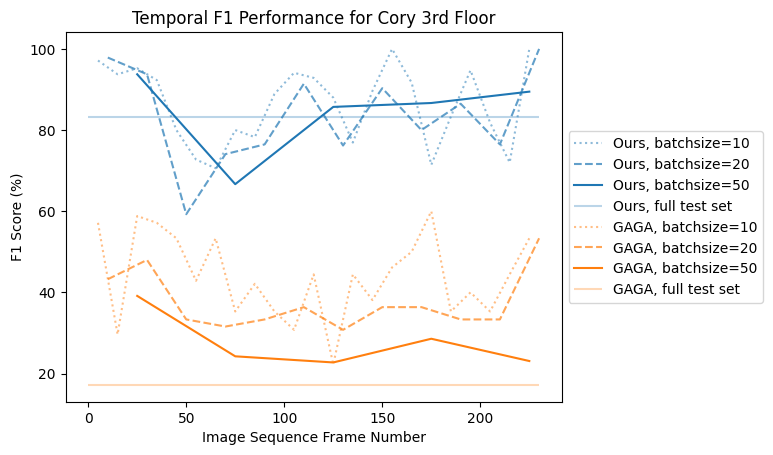}
        \caption{Plot of temporal F1 performance for the Cory 3rd Floor dataset}
        \label{fig:results-temporal-cory3rd}
    \end{subfigure}
    \\
    \begin{subfigure}{\linewidth}
        \centering
        \includegraphics[width=0.9\linewidth]{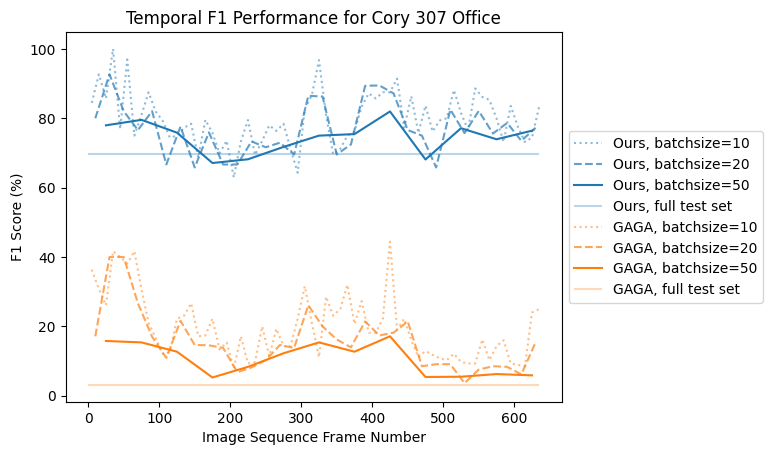}
        \caption{Plot of temporal F1 performance for the Cory 307 Office dataset}
        \label{fig:results-temporal-cory307}
    \end{subfigure}
    \caption{Plots of temporal F1 performance comparing our mask association method against GAGA's \cite{lyu2024gaga}, computed for image batch sizes of 10, 20, and 50. F1 Scores computed from the full test set, reported in \cref{tab:results_maskassociation}, are plotted as horizontal lines.}
    \label{fig:results-temporal}
\end{figure}
\noindent \textbf{2D Mask Association:}
\label{results-2dmaskassoc}
We compare our mask association approach against GAGA \cite{lyu2024gaga}, which serves as our baseline. GAGA relabels 2D masks with multi-view consistent object IDs using a 3D-aware memory bank, and uses these relabeled masks to train an identity encoding for each Gaussian. For evaluation, we compare our method against GAGA's intermediate relabeled outputs to assess mask association performance.
\Cref{tab:results_maskassociation} reports quantitative results on both datasets, demonstrating that our method significantly outperforms GAGA's mask association across all metrics, while achieving a 2-4x speedup in pipeline runtime. The right side of \cref{fig:results-maskassoc-cory307} presents qualitative mask association results on the Cory 307 Office dataset.

GAGA exhibits a pronounced tendency to over-merge objects, assigning the same ID to distinct objects\textemdash for example, in \cref{fig:results-maskassoc-cory307}, where most masks in GAGA's output possess the same light blue color.
This over-merging behavior persists even when GAGA performs mask association using full-image segmentations produced by SAM's \textit{segment-everything} mode rather than masks restricted to target object classes; as illustrated in \cref{fig:results-gaga-fullseg}, objects in earlier frames are correctly separated whereas later frames show severe over-merging, with nearly all masks assigned the same ID, visualized by the purple color.
In contrast, our method accurately distinguishes different objects while also maintaining consistent IDs for a given object across multiple views, as shown in \cref{fig:results-maskassoc-cory307}.
GAGA, on the other hand, struggles to maintain multi-view mask consistency: object colors fluctuate across viewpoints\textemdash for example, the \textit{table} object in \cref{fig:results-maskassoc-cory307} changes from pink to light blue to green and back to light blue.

To further analyze GAGA's over-merging behavior, we compute F1 scores over progressively larger image batches, e.g. 10, 20, and 50 images, and plot their evolution over the course of the image sequence in \cref{fig:results-temporal}. 
Qualitatively, we observe that GAGA associates object masks correctly in viewpoints early in the image sequence when there are relatively few objects present in its 3D-aware memory bank.
However, as additional objects are added to the memory bank, GAGA increasingly over-merges objects, leading to multiple distinct objects collapsing to a single ID. This can cause different objects observed early and late in the image sequence to be assigned the same identifier. As a result, F1 scores computed on small batches fail to capture this degradation, and are therefore artificially inflated.
We observe that as the batch size increases from 10 to 20 to 50 images, GAGA's overall performance deteriorates, as reflected by the consistently lower F1 curves associated with larger batch sizes in \cref{fig:results-temporal}.
This behavior explains why GAGA's F1 scores computed over the full test set, reported in \cref{tab:results_maskassociation}, are substantially lower than the per-batch scores shown in the temporal plots.

GAGA's over-merging issue is further evidenced by \cref{tab:results_numberofmasks}, which reports a significantly smaller number of unique masks compared to ground truth for both datasets, indicating frequent erroneous mask associations of distinct objects.
Our method on the other hand exhibits stable temporal F1-score performance across batch sizes and consistently outperforms GAGA, as shown in \cref{fig:results-temporal}. Moreover, our approach produces a number of unique masks closely matching ground truth for each dataset, as indicated in \cref{tab:results_numberofmasks}. 

We attribute our superior performance to two key factors: our enhanced depth-processing method for associating 2D masks with 3D Gaussians, denoted as step~2A in \cref{fig:pipelineoverview}, as well as the incorporation of semantic labels during mask merging in object codebook construction, corresponding to step~2B. The impact of these components is further substantiated by the pipeline component ablation studies presented in \cref{sec:ablations_pipeline}.
Together, these improvements mitigate erroneous object merges. Without semantic constraints, GAGA risks erroneously merging spatially proximate but semantically distinct sets of Gaussians. Moreover, GAGA's depth processing strategy tends to excessively include spurious background Gaussians, as illustrated in \cref{fig:depthgaga}, which further exacerbates incorrect merging, leading to the observed over-merging behavior.
\begin{table}
    \centering
    \caption{Object detection evaluation, comparing our results and baseline OWLv2 \cite{minderer2023owlv2} detections.}
    \label{tab:results_objectdetection}
    \footnotesize
    \begin{tabular}{l *4{S[table-format=2.2]}}
    \toprule
    & \multicolumn{2}{c}{Cory 3rd Floor} & \multicolumn{2}{c}{Cory 307 Office} \\
    \cmidrule(lr){2-3} \cmidrule(lr){4-5}
    Method & {mAP $\uparrow$} & {mLAMR $\downarrow$} & {mAP $\uparrow$} & {mLAMR $\downarrow$} \\
    \midrule
    Ours & \textbf{41.78} & \textbf{69.77} & \textbf{41.62} & \textbf{63.92} \\
    OWLv2 \cite{minderer2023owlv2} & 28.15 & 78.30 & 33.27 & 73.19 \\
    \bottomrule
    \end{tabular}
\end{table}
% Bbox -- Cory 3rd Floor
\begin{figure}
    \centering
    \begin{subfigure}[t]{0.4\linewidth}
        \includegraphics[width=\linewidth]{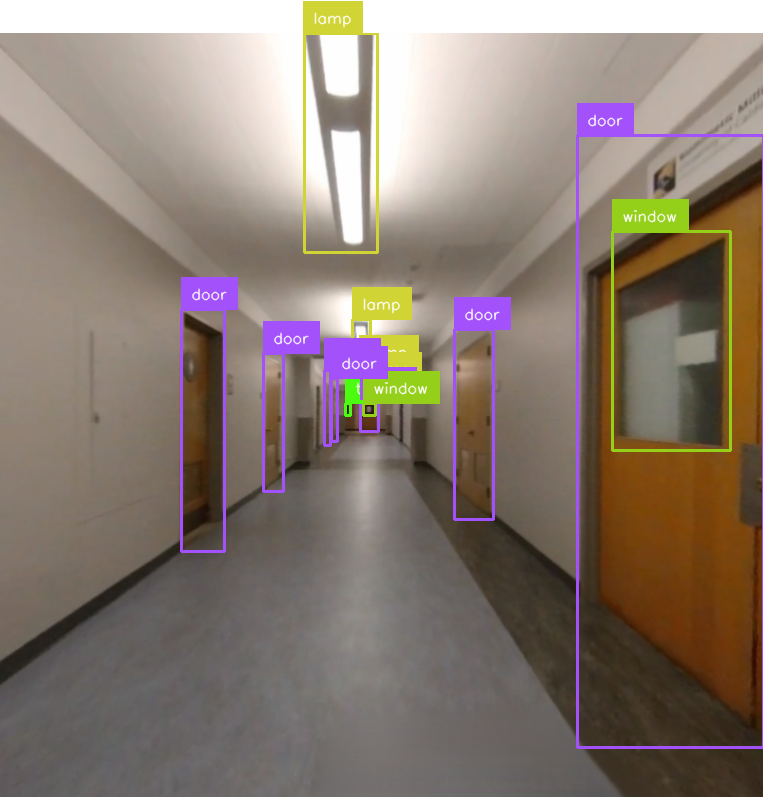}
        \caption{Ground Truth}
    \end{subfigure} %
    \begin{subfigure}[t]{0.4\linewidth}
        \includegraphics[width=\linewidth]{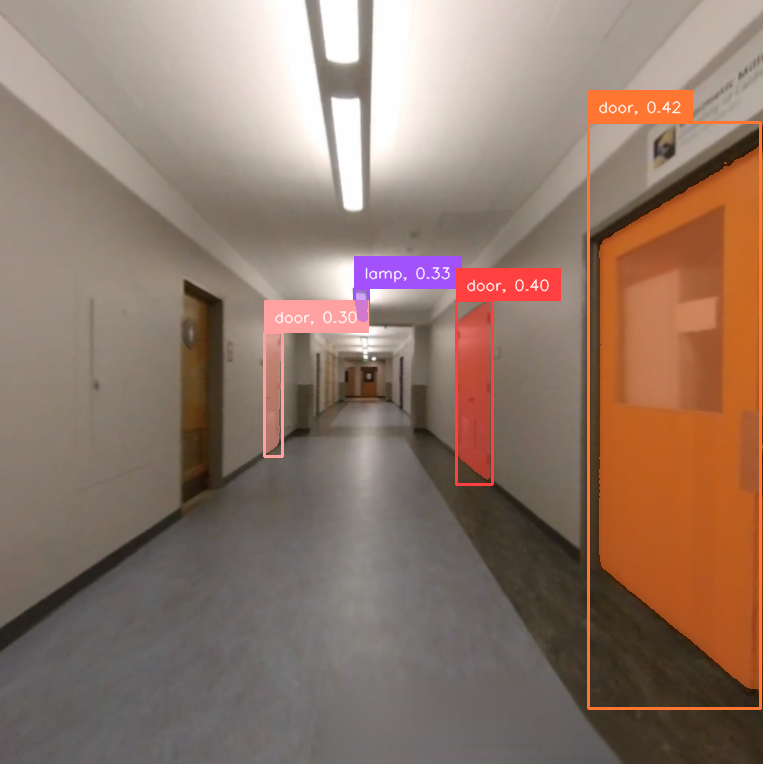}
        \caption{OWLv2 \cite{minderer2023owlv2}}
    \end{subfigure}
    \\
    \begin{subfigure}[t]{0.4\linewidth}
        \includegraphics[width=\linewidth]{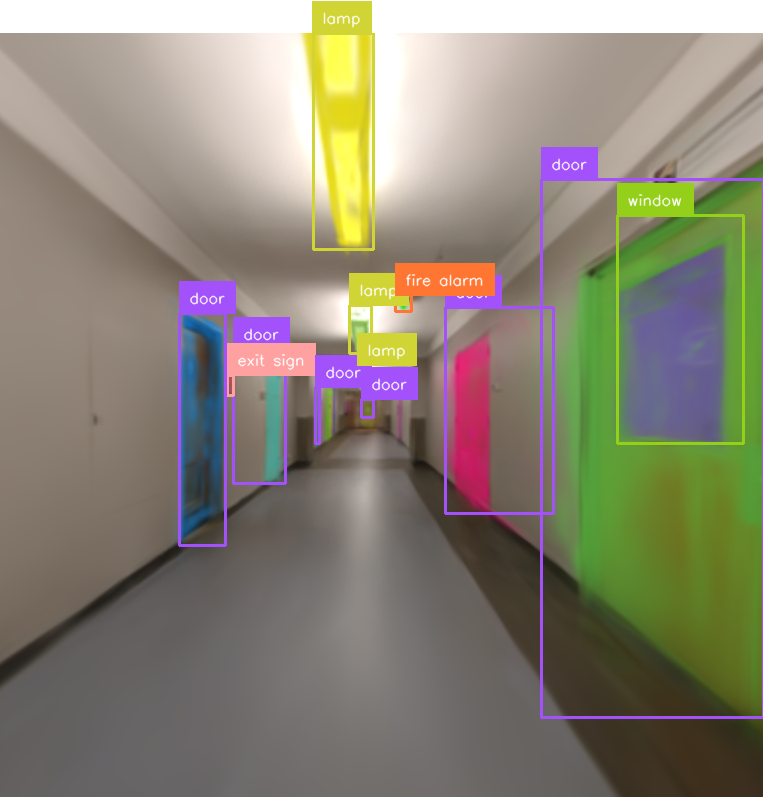}
        \caption{Ours}
    \end{subfigure} %
    \begin{subfigure}[t]{0.4\linewidth}
        \includegraphics[width=\linewidth]{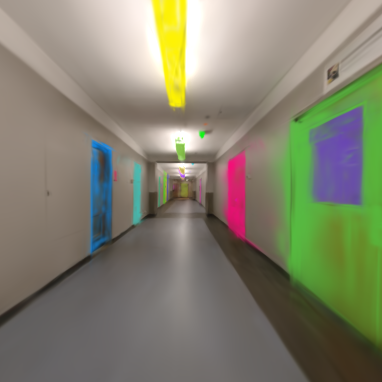}
        \caption{Ours w/o annotations}
    \end{subfigure}
    \caption{Qualitative results of object detection for the Cory 3rd Floor dataset.}
    \label{fig:results-bbox-cory3rd-2}
\end{figure}
% ABLATION TABLE-----
\begin{table*}
\centering
\caption{Ablation study for pipeline components. Cells are colored from best (green) to worst (red) per column.}
\label{tab:ablation_pipeline}
\footnotesize
\begin{subtable}{\linewidth}
    \centering
    \begin{tabular}{@{} l *{12}{S[table-format=2.2]}}
    \toprule
    & \multicolumn{6}{c}{Cory 3rd Floor} & \multicolumn{6}{c}{Cory 307 Office} \\
    \cmidrule(lr){2-7} \cmidrule(lr){8-13}
    & \multicolumn{4}{c}{2D Mask Association} & \multicolumn{2}{c}{Object Detection} & \multicolumn{4}{c}{2D Mask Association} & \multicolumn{2}{c}{Object Detection}\\
    \cmidrule(lr){2-5} \cmidrule(lr){6-7} \cmidrule(lr){8-11} \cmidrule(lr){12-13}
    Pipeline & 
    {mIoU} & {Prec.} & {Rec.} & {F1} &
    {mAP}  & {mLAMR $\downarrow$} &
    {mIoU} & {Prec.} & {Rec.} & {F1} &
    {mAP}  & {mLAMR $\downarrow$}\\
    \midrule
    \textbf{Full} & 
    \cellcolor{rank1}75.84 & \cellcolor{rank1}82.61 & 
    \cellcolor{rank1}84.07 & \cellcolor{rank1}83.33 & 
    \cellcolor{rank1}41.78 & \cellcolor{rank1}69.77 &
    \cellcolor{rank2}66.01 & \cellcolor{rank3}67.05 & 
    \cellcolor{rank2}72.54 & \cellcolor{rank1}69.69 & 
    \cellcolor{rank4}41.62 & \cellcolor{rank2}63.92 \\
    w/o depth processing & 
    \cellcolor{rank6}48.85 & \cellcolor{rank6}79.22 & 
    \cellcolor{rank6}53.98 & \cellcolor{rank6}64.21 & 
    \cellcolor{rank6}15.07 & \cellcolor{rank6}89.78 &
    \cellcolor{rank7}52.36 & \cellcolor{rank2}67.49 & 
    \cellcolor{rank7}56.15 & \cellcolor{rank7}61.30 & 
    \cellcolor{rank8}9.69 & \cellcolor{rank8}88.87 \\
    w/o semantic constraint & 
    \cellcolor{rank5}72.36 & \cellcolor{rank4}81.82 & 
    \cellcolor{rank5}79.65 & \cellcolor{rank5}80.72 & 
    \cellcolor{rank5}37.71 & \cellcolor{rank5}72.00 &
    \cellcolor{rank8}47.47 & \cellcolor{rank1}70.79 & 
    \cellcolor{rank8}51.64 & \cellcolor{rank8}59.72 & 
    \cellcolor{rank7}28.08 & \cellcolor{rank7}75.46 \\
    w/o $1^\text{st}$ filtering & 
    \cellcolor{rank1}75.84 & \cellcolor{rank1}82.61 & 
    \cellcolor{rank1}84.07 & \cellcolor{rank1}83.33 & 
    \cellcolor{rank4}39.41 & \cellcolor{rank4}71.94 &
    \cellcolor{rank5}65.61 & \cellcolor{rank6}66.92 & 
    \cellcolor{rank5}71.31 & \cellcolor{rank4}69.05 & 
    \cellcolor{rank3}41.95 & \cellcolor{rank1}63.46 \\
    w/o spatial merging & 
    \cellcolor{rank4}75.45 & \cellcolor{rank5}81.20 & 
    \cellcolor{rank1}84.07 & \cellcolor{rank4}82.61 & 
    \cellcolor{rank3}41.58 & \cellcolor{rank3}69.95 &
    \cellcolor{rank6}64.06 & \cellcolor{rank7}57.72 & 
    \cellcolor{rank6}70.49 & \cellcolor{rank5}63.47 & 
    \cellcolor{rank1}43.44 & \cellcolor{rank4}66.47 \\
    w/o $2^\text{nd}$ filtering & 
    \cellcolor{rank1}75.84 & \cellcolor{rank1}82.61 & 
    \cellcolor{rank1}84.07 & \cellcolor{rank1}83.33 & 
    \cellcolor{rank1}41.78 & \cellcolor{rank1}69.77 &
    \cellcolor{rank2}66.01 & \cellcolor{rank3}67.05 & 
    \cellcolor{rank2}72.54 & \cellcolor{rank1}69.69 & 
    \cellcolor{rank5}40.97 & \cellcolor{rank3}64.38 \\
    w/o object filtering & &&&&&&
    \cellcolor{rank1}70.55 & \cellcolor{rank8}52.11 & 
    \cellcolor{rank1}75.82 & \cellcolor{rank6}61.77 & 
    \cellcolor{rank2}42.15 & \cellcolor{rank5}67.43 \\
    w/o outlier removal &  &&&&&&
    \cellcolor{rank2}66.01 & \cellcolor{rank3}67.05 & 
    \cellcolor{rank2}72.54 & \cellcolor{rank1}69.69 & 
    \cellcolor{rank6}30.14 & \cellcolor{rank6}75.07 \\
    \bottomrule
\end{tabular}
\end{subtable}
\end{table*}
%-------------------

\noindent \textbf{Object Detection:}
\label{sec:results-objdetection}
We evaluate object detection quality by comparing predicted bounding boxes against ground-truth annotations, using OWLv2~\cite{minderer2023owlv2} detections as a baseline.
Our codebook is able to aggregate detections across multiple viewpoints, thereby recovering missed detections by OWLv2 in individual viewpoints and improving robustness against erroneous but infrequent detections.
As shown in \cref{tab:results_objectdetection}, our approach outperforms the baseline across all evaluated metrics, demonstrating the benefit of multi-view aggregation over single-view detection.

\Cref{fig:results-bbox-cory3rd-2} demonstrates the effectiveness of our approach\textemdash bounding boxes generally tightly enclose the corresponding objects, indicating accurate localization. However, some false positives and missed detections persist due to OWLv2 errors. Duplicate objects occasionally arise from floating Gaussians, particularly around transparent or reflective surfaces, which can skew spatial overlap and hinder merging.
In more challenging environments such as the Cory 307 Office dataset, our approach maintains strong performance despite higher object density and class diversity, successfully aggregating detections missed by OWLv2.
Many of the errors, including over-extended bounding boxes, duplicate instances, and occasional mislabeling, are largely attributable to persistent floaters and inconsistencies in OWLv2 detections across viewpoints.
%==================== ABLATIONS FOR PIPELINE STAGES ==============
\subsection{Ablation Studies}
\label{sec:ablations}
\noindent \textbf{Ablation Study for Thresholds:} \label{sec:ablationthresholds}
To determine optimal threshold values for each pipeline step, we conduct a staged ablation study in which pipeline components are introduced incrementally. At each stage, we sweep the threshold of the newly added step while keeping previously selected thresholds fixed. This process is repeated until all pipeline component thresholds are determined. The final selected values are summarized in \cref{tab:thresholds} and are selected through joint evaluation on both datasets to support generalization across indoor environments.

\noindent \textbf{Ablation Study for Pipeline Steps:}
\label{sec:ablations_pipeline}
We further conduct an ablation study to evaluate the contribution of each stage in our pipeline, specifically steps 2A-E and 3A-B in \cref{fig:pipelineoverview}. Results on both datasets are reported in \cref{tab:ablation_pipeline}.

Across both datasets, depth-based processing and semantic-constraints during merging are critical components of our pipeline. Removing either leads to significant performance degradation in both mask association and object detection, for simple as well as complex scenes.

On the Cory 3rd Floor dataset, the full pipeline achieves the best performance across all metrics. Ablating the second low-weight Gaussian filtering stage yields similar results, which is expected since Gaussian filtering does not directly affect mask association. In relatively simple scenes, earlier stages may already produce sufficiently clean segmentations, limiting the benefit of further geometric refinement for object detection.

For the more challenging Cory 307 Office dataset, the complete pipeline again achieves the best overall performance.
Ablating the first low-weight Gaussian filtering step or spatial merging step yields similar object detection performance but worse mask association, indicating their importance for maintaining multi-view mask consistency.
Including the second low-weight Gaussian filtering stage provides a modest improvement in object detection performance.
Removing object filtering substantially reduces mask association precision and increases mLAMR, and finally, ablating spatial outlier removal markedly degrades object detection performance, highlighting its role in eliminating residual floater Gaussians and improving spatial coherence.

\section{Conclusion}
In this paper, we presented a pipeline for associating 2D masks across different viewpoints into a coherent 3D object codebook, enabling 3D object detection and segmentation within 3D Gaussian Splatting.
Through mask-association and object detection evaluations, we demonstrate significant performance improvements over baseline methods. Qualitative results also illustrate improved object completeness and spatial coherence, even in dense and occluded environments.

{
    \small
    \bibliographystyle{ieeenat_fullname}
    \bibliography{main}
}


\end{document}